\documentclass[12pt,reqno,a4paper]{amsart}

\usepackage{latexsym}
\usepackage{amsmath}
\usepackage{amscd}
\usepackage{amssymb}
\usepackage{mathrsfs}
\usepackage{mathtools}
\usepackage{comment}
\usepackage{color,framed}
\usepackage{enumerate}
\usepackage{soul}
\usepackage{stmaryrd}
\usepackage[ruled]{algorithm2e} 
\usepackage{braket}

\usepackage{latexsym}
\usepackage{amsmath}
\usepackage{amscd}
\usepackage{amssymb}
\usepackage{mathrsfs}
\usepackage{mathtools}
\usepackage{comment}
\usepackage{color,framed}
\usepackage{enumerate}
\usepackage{soul}
\usepackage{stmaryrd}
\usepackage[ruled]{algorithm2e} 
\usepackage{braket}

\usepackage{graphicx}
\usepackage{subfigure}
\usepackage[colorlinks]{hyperref}
\usepackage{url}

\newtheorem{theorem}{Theorem}[section]

\newtheorem{proposition}[theorem]{Proposition}
\newtheorem{remark}[theorem]{Remark}

\def\RR{{\mathbb R}}
\newcommand{\pair}[1]{\left<#1\right>}
\DeclareMathOperator{\divergence}{div}
\DeclareMathOperator{\Diff}{Diff}
\DeclareMathOperator{\ad}{ad}
\DeclareMathOperator{\argmin}{argmin}
\DeclareMathOperator{\argmax}{argmax}

\numberwithin{equation}{section}

\begin{document}

\title{String Methods for Stochastic Image and Shape Matching} 

\author[A. Arnaudon]{Alexis Arnaudon}
\author[D. Holm]{Darryl D Holm}
\author[S. Sommer]{Stefan Sommer}

\address{AA, DH: Department of Mathematics, Imperial College, London SW7 2AZ, UK}
\address{SS: Department of Computer Science (DIKU), University of Copenhagen,
  DK-2100 Copenhagen E, Denmark}

\maketitle

\begin{abstract}
Matching of images and analysis of shape differences is traditionally pursued by energy minimization of paths of deformations acting to match the shape objects.  In the Large Deformation Diffeomorphic Metric Mapping (LDDMM) framework, iterative gradient descents on the matching functional lead to matching algorithms informally known as Beg algorithms.  When stochasticity is introduced to model stochastic variability of shapes and to provide more realistic models of observed shape data, the corresponding matching problem can be solved with a stochastic Beg algorithm, similar to the finite temperature string method used in rare event sampling. In this paper, we apply a stochastic model compatible with the geometry of the LDDMM framework to obtain a stochastic model of images and we derive the stochastic version of the Beg algorithm which we compare with the string method and an expectation-maximization optimization of posterior likelihoods. The algorithm and its use for statistical inference is tested on stochastic LDDMM landmarks and images.


\end{abstract}


\section{Introduction}
Image and shape variations are often modelled by the action of the diffeomorphism
group on the data space. This approach is the basis for the Large Deformation
Diffeomorphic Metric Mapping (LDDMM) method that provides a general framework for
representing and analysing variations of various types of shape data - images, 
landmarks, curves, surfaces and tensor fields
- through a right-invariant metric structure on the 
diffeomorphism group. In recent works
\cite{arnaudon_geometric_2018,arnaudon2016stochastic2},
a general framework for modelling stochastic shape variability has been introduced
based on right-invariant perturbation of the shape evolution.
In this paper, we specialize the general framework to obtain a model
for stochastic shape variation in images.
The introduction of the noise and the derivation of the corresponding stochastic
evolution equations build strongly upon the momentum representation of images of 
\cite{bruveris2011momentum}.
In this work, we outline the theoretical background, flow equations, and matching algorithms which 
we will use to derive stochastic versions of the deterministic matching 
algorithms between images or other shape data. 
This will be a stochastic generalisation of the gradient-based minimization algorithm known as the Beg
algorithm, arguably the most fundamental LDDMM matching algorithm \cite{beg2005computing}.
We will show how the resulting iterative minimization scheme is analogous to 
string methods as used in rare event sampling
\cite{e_string_2002,e_finite_2005}, and how the scheme
relates to a stochastic approximation Expectation-Maximization algorithm
\cite{dempster_maximum_1977} for inference of 
optimal trajectories with noise drawing links to the estimation of principal
curves in statistics \cite{hastie_principal_1989,tibshirani_principal_1992}.
While the string method needs a fully convergent optimization for
each noise realization, the finite temperature string method takes only
one gradient descent step per realized noise. This distinguishes the finite temperature
string method as a computationally efficient algorithmic tool for statistical
inference on high-dimensional shape spaces.
\begin{figure}[htpb]
\centering
\includegraphics[width = 0.8\textwidth, trim = 0 80 0 80, clip]{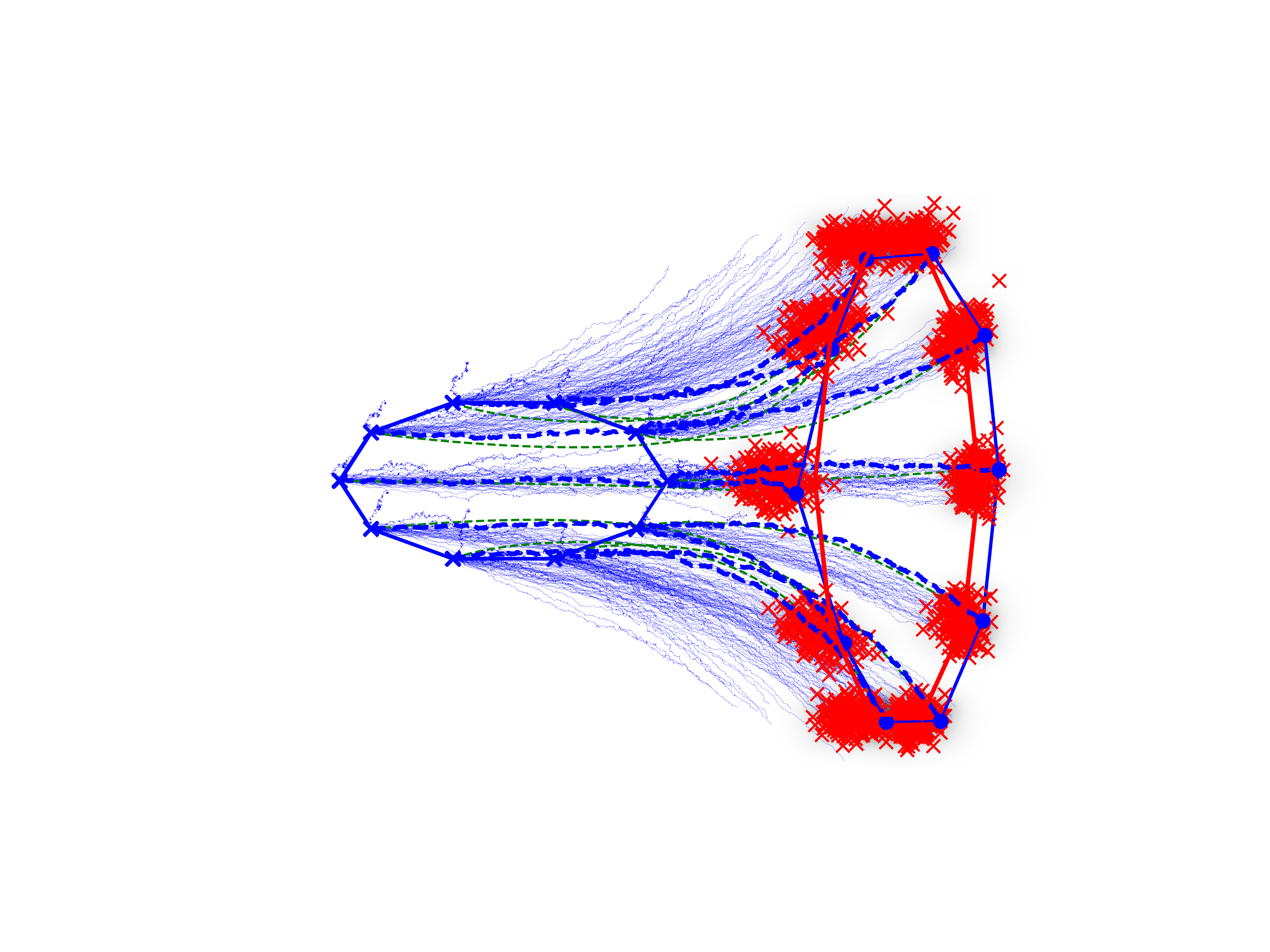}
\caption{Stochastic landmark strings matching an initial configuration $I_0$ (blue solid
  lines/crosses) with a target configuration $I_1$ (red solid line). Samples from 
  the endpoint distribution (red crosses) are shown together with a subset of the strings. The
finite energy mean string (fat blue dashed) appears as an unperturbed solution
as the matching algorithm converges. The figure shows inexact matching, which is
the focus problem of the paper, as can be seen by the
non-zero variance of the endpoint distribution around the target.
}
\label{fig:string1}
\end{figure}

Images are in the LDDMM model matched through the push-forward
action $g.I=I\circ g^{-1}$ of a diffeomorphism
$g\in\Diff(\Omega)$ on an image $I:\Omega\rightarrow\RR$ defined on a domain
$\Omega\subseteq\RR^d$. The
corresponding optimal deformation flows are governed by the Euler-Poincar\'e equation on the diffeomorphism group (EPDiff)
which, in vector notation, takes the form
\begin{align}
  \partial_t m_t
  +( u_t\cdot \nabla)  m_t
  +m_t\cdot (D u_t)^T
  +\divergence(u_t)m_t
  =
  0\, , 
  \label{det-vector-EPDiff-intro}
\end{align}
together with the reconstruction equation
$
    \partial_t g_t 
    = 
    u_t\circ g_t
    $
whose solution determines the corresponding diffeomorphisms, or warps, $g\in\Diff(\Omega)$.
Here, $u_t$ is a vector field on the image and $m_t$ its dual momentum field. $D$ is the spatial Jacobian matrix, $\nabla$ the gradient, and $\divergence$ the divergence.
The solutions of these equations are geodesic motions on the group of diffeomorphisms, 
a property which is crucial for image matching, provided a proper class of Lagrangians is chosen. 
For details, see, e.g., \cite{younes_shapes_2010}. 

We introduce stochasticity by adding a perturbation term to the 
reconstruction equation in a way that preserves the momentum
map \cite{bruveris2011momentum}. As a result, we will arrive at the stochastic version of the image EPDiff equation that in vector form
generalizes the deterministic equation with the addition of $J$ 
Eulerian fields $\sigma_l\in\mathfrak X(\Omega)$ multiplied by Stratonovich
noise
\begin{align}
  \mathrm{EPDiff\,\eqref{det-vector-EPDiff-intro}}
  +
  \sum_{l=1}^J
  \big(
  (\sigma_l\cdot \nabla) m_t
  +m_t\cdot (D\sigma_l)^T
  +\divergence(\sigma_l)m_t
  \big)
  \circ
  dW^l_t
  =
  0
  \,. 
  \label{sto-vector-EPDiff-intro}
\end{align}

The abstract form of this equation was presented in
\cite{arnaudon_geometric_2018} building on the derivation of
\cite{holm2015variational}. In fluid dynamics, 
the stochastic EPDiff equation was derived in vector form in 
\cite{HoTy2016} in a variational setting corresponding to exact
matching. Here, we derive the stochastic image EPDiff equation in
vector notation to make clear its role in extending the common vector form
\eqref{det-vector-EPDiff-intro} of the deterministic EPDiff equation, and
we focus on the case of inexact matching.

Moreover, because the Beg matching algorithm optimizes to fulfil the
momentum equation, a flow equation expressed with the momentum map, the preservation of the momentum map
allows derivation of a stochastic counterpart of the algorithm with equivalent
structure. For fixed noise realization, we will see that the algorithm directly
extends the Beg algorithm; and for variable noise realization, it has a direct
counterpart in the string method \cite{e_string_2002}. Namely, upon changing the noise for
each iteration of the algorithm, the algorithm becomes a shape counterpart to the
finite temperature string method of \cite{e_finite_2005}.

The introduction of stochasticity to model non-deterministic dynamics 
results in statistical models of image and shape data from 
distributions of data observed at fixed points in time. This leads to
geometrically intrinsically defined probability distributions on the nonlinear
shape spaces and allows for quantifying the uncertainty, quality, and robustness of
matchings of pairs of data points. The approach, in addition, suggests that 
statistical inference can be based on
parameter estimation in the model of parameters such as initial conditions of the flows and noise 
structure. As a particular case of this, we use string methods as a
computationally efficient tool for 
inferring a version of the Frech\'et mean with noise.

\subsection{Paper Outline}

In Section \ref{sec:background}, we survey the
geometric framework behind the general stochastic shape model and its
introduction through the momentum equation. In Section \ref{sec:sto-images},
we apply the model to images to derive the stochastic image Euler-Poincar\'e
equation \eqref{sto-vector-EPDiff-intro} in vector notation in the case of
inexact matching.
We proceed in Section \ref{sec:sto-inexact-match} with algorithms for stochastic inexact matching,
description of the string methods and their relation to the
Beg algorithm and expectation-maximisation (EM) estimation of maximum a posteriori curves.
Statistical aspects of the model will be discussed in Section
\ref{sec:stat}, and we perform numerical experiments on landmarks data in Section 
\ref{sec:experiments} before giving concluding remarks.

\section{Deterministic and Stochastic LDDMM Shape Analysis} \label{sec:background}

We here review the geometric framework for stochastic dynamics of general shapes as
presented in \cite{arnaudon_geometric_2018,arnaudon2016stochastic2}. The model
is based on 
parametric stochastic deformations in fluid dynamics introduced in
\cite{holm2015variational} and stochastic coadjoint motion
\cite{arnaudon2016noise} in finite dimensional Lie groups. The preservation of the geometrical structure of
LDDMM when passing to the stochastic setting is obtained by introducing noise
that preserves the momentum map \cite{HoMa2004} and thereby the momentum
map representation of images and shapes \cite{bruveris2011momentum}. We describe
the deterministic LDDMM construction to the extent necessary for providing context for the
derivation of the stochastic dynamics.
\begin{remark}
  Stochastic evolution of shapes has been considered in the literature earlier 
  \cite{TrVi2012,vialard2013extension} and more recently with stochastic
  landmark dynamics in \cite{marsland2016langevin}. Both approaches add
  stochasticity only in the momentum equation of the dynamics.
  The present model introduces noise that preserves the original geometrical structure of the deterministic equations. 
  As a consequence, the solutions remain diffeomorphisms with a controlled spatial correlation of the noise. 
  As demonstrated in the introduction of \cite{arnaudon_geometric_2018}, the limit of large number of landmarks retains the original spatial correlation of the noise. 
  This is important as the particular shape feature, e.g. the number of landmarks, can be a modelling choice while the spatial correlation can be an intrinsic property of the shape or image, and, in the landmark case, independent on the chosen number of landmarks. 
  The structure of the noise should thus be the same if inferred using a small or large number of landmarks, or even shapes or complete images. 

\end{remark}

\subsection{Large Deformation Inexact Matching}

In the deterministic setting, shape matching is in the LDDMM framework defined
from the energy functional
\begin{align}
    E(u;I_0,I_1) = \int_0^1 l( u_t) dt + \frac{1}{2\lambda^2} \|g_1.I_0 - I_1\|^2\, ,
    \label{E-functional}
\end{align}
over time dependent vector fields $u_t\in\mathfrak X(\Omega)$ for some domain
$\Omega\subseteq\RR^d$ and with weight $\lambda\in\mathbb R^+$. We often write
$E(u)$ making the dependence on $I_0,I_1$ implicit. The
rightmost term of the energy is a dissimilarity measure between the shape
$I_1$ and the shape $I_0$ that is transformed by the action of a diffeomorphism,
or warp, $g_1\in\Diff(\Omega)$. The left-most term of
the energy is a Lagrangian on the flow $u_t$, taken to by hyper-regular on a subspace of $\mathfrak X(\Omega)$ so that the associated momentum variable is well-defined via the Legendre transformation. The final diffeomorphism $g_1$
that acts to deform $I_0$ is obtained from the reconstruction equation
\begin{align}
  \partial_t g_t = u_t \circ g_t\, ,\quad g_0=\mathrm{Id}_{\Omega}\, , 
    \label{rec-det}
\end{align}
evaluated at $t=1$. Minimizing \eqref{E-functional} for $\lambda<\infty$ is
called \emph{inexact} matching since the dissimilarity term will generally be
non-vanishing at minimal $u_t$. In cases where $I_1$ lies in the orbit of
$\Diff(\Omega)$ acting on $I_1$, one can instead require exact matching
corresponding to $\lambda=0$ and the dissimilarity term being zero at
optimal $u_t$. In this case, the system is solved as a hard constraint on the solution, via a shooting method for example.  
For images, the orbit criterion is seldom
satisfied in practice which leads to the inexact matching case being used in
general. Even for shape structures such as landmarks where the action is
transitive, the presence of noise in observed data strongly suggests using inexact
matching to avoid improbable warps.

We will assume here that shapes are elements of a vector space $V$ on which 
$\Diff(\Omega)$ acts.
This vector space is assumed to have a scalar product, or pairing to be able to define its dual vector space $V^*$. 
When the shapes are
images, i.e. $I_0,I_1:\Omega\rightarrow\RR$, the action is by 
push-forward $g.I=I\circ g^{-1}$. In the case of $n$ landmarks in $\Omega$,
$I=(\mathbf q_1,\ldots,\mathbf q_n)$, the action is by evaluation of $g$ on the landmarks, i.e. 
$g.I=(g(\mathbf q_1),\ldots,g(\mathbf q_n))$. For shapes such as curves or surfaces, the action
is defined analogously though in this case, the vector space assumption is not
satisfied. However, the construction can be generalized to cover such shape spaces as well.

\subsection{The Momentum Representation of Shapes and Images}

Optimal vector fields $u_t$ for \eqref{E-functional} satisfy the
condition $\nabla_{u}E(u)=0$, which corresponds to the directional derivative along $u= (u_t)_{t\in [0,1]}$, and therefore the corresponding Euler-Poincar\'e equation (or EPDiff equation \eqref{det-vector-EPDiff-intro})
\begin{align}
    \partial_t\frac{\delta l}{\delta u} + \ad^*_{u_t} \frac{\delta l}{\delta u}=0\,, 
    \label{EP}
\end{align}
where $\mathrm{ad}^*: \mathfrak g \times \mathfrak g^* \to \mathfrak g^*$ is the coadjoint action of the Lie algebra on its dual and $\frac{\delta }{\delta u}$ are variational derivative of the functional $l(u)$ with respect to the functions $u(x)$ on the domain $\Omega$. 
Both equations can be understood in terms of momentum maps as commonly used in
geometric mechanics \cite{holm_geometric_2011} and as used in the momentum map
representation of images and shapes \cite{bruveris2011momentum}. We here briefly
outline the construction.

The space of vector fields $\mathfrak X(\Omega)$ can be considered the Lie
algebra $\mathfrak g$ of $G=\Diff(\Omega)$, and
the Lagrangian $l$ maps $u\in\mathfrak X(\Omega)$ to elements 
$m=\frac{\delta l}{\delta u}$ of the dual of the Lie algebra $\mathfrak g^*$
giving 1-form densities with the pairing 
$\pair{\xi,u}=\int_\Omega \xi(\mathbf x)(u(\mathbf x))$, $\xi\in\mathfrak g^*$. 
The Lagrangian $l$ is often defined from an
inner product $l(u)=|u|_l^2=\left<u,Lu\right>_{L^2}$ using a positive,
self-adjoint operator $L$, in which case $\frac{\delta l}{\delta u}=Lu$.
In the sequel, we will denote by $K$, the Green's function of the operator $L$. 

For critical points $u_t$ of \eqref{E-functional}, the momentum takes
a particular form coming from the cotangent-lift momentum map. In this
setting, the map is denoted $\diamond:V\times V^*\rightarrow\mathfrak{g}^*$ with domain 
identified with the cotangent bundle $TV^*$. The momentum map is defined from the
infinitesimal action $u.I\in V$ of $u\in\mathfrak{g}$ on shapes $I$ arising from the
action of $\Diff(\Omega)$ on $I$: If $\partial_t|_{t=0}\phi_t=u$ then
$u.I=\partial_t|_{t=0}\phi_t.I$. A covector $f\in
V^*$ can be paired with $u.I$ and the map $\diamond$ is defined by evaluation on
$u\in\mathfrak{g}$ by
\begin{align}
    \langle I\diamond f, u\rangle_{\mathfrak g^* \times \mathfrak g}:= \langle f, u.I\rangle_{V^*\times V}\, . 
\end{align}
Elements of the dual space $V^*$ can be represented by vectors in $V$ using the
$L^2$-pairing $\pair{f,I}=\int_\Omega f(\mathbf x)I(\mathbf x)dx$, $f\in V^*$ which in
turn defines the flat map $\flat:V\rightarrow V^*$. 
It is shown in \cite{bruveris2011momentum} that $u_t$ is critical for
\eqref{E-functional} in the sense $\nabla_uE(u)=0$ if and only if
\begin{align}
    \frac{\delta l}{\delta u_t}
    = - \frac{1}{\lambda^2} J_t^0\diamond ( g_{t,1}.(J_1^0-J_1^1)^\flat )\, ,
    \label{momentum-eq}
\end{align}
where $g_{t,s}$ denotes the solution of the reconstruction equation at time $t$
started at $s$, $J_t^0=g_{t,0}.I_0$ is the shape $I_0$ flowed forward to time
$t$, and $J_t^1=g_{t,1}.I_1$ is the shape $I_1$ flowed backward from $s=1$ to
$t$. The momentum is thus constrained by the momentum map applied to the
transported shapes using the value at time $t$ of the diffeomorphism $g$.

The EPDiff equation \eqref{EP} can now be derived from \eqref{momentum-eq} using the fact that
the cotangent-lift momentum map is infinitesimally equivariant and taking the time
derivative of the momentum $m$. The only formal difference in the derivation between different
shape types is the particular form of the infinitesimal action of $u\in\mathfrak
g$ on $I$. For images, $u.I=-\nabla I\cdot u$ resulting in the momentum map
\begin{align}
  \left \langle I
  \diamond
  f,u\right \rangle
  =
  \int_\Omega
  -(\nabla I\cdot u)f
  d\mathbf x
   \, .
  \label{momentum-map-images}
\end{align}
and momentum equation
\begin{align}
  Lu_t
  =
  -
  \frac{1}{\lambda^2} 
  \left|\det Dg_{t,1}^{-1}\right| (J_t^0-J_t^1)\nabla J_t^0
  \, ,
  \label{momentum-equation-images}
\end{align}
where $Dg_{t,1}^{-1}$ stands for the Jacobian of the inverse map $g_{t,1}^{-1}$. 
For landmarks,
the infinitesimal action is $u.I=(u(\mathbf q_1),\ldots,u(\mathbf q_N))$ and the 
momentum map becomes 
\begin{align}
  (\mathbf q^1,\ldots,\mathbf q^n)
  \diamond
  (\mathbf p^1,\ldots,\mathbf p^n)^\flat
  =
  \sum_{i=1}^n\mathbf p_i\delta_{\mathbf q_i}\,, 
  \label{momentum-map-landmarks}
\end{align}
which, for matching $I_0=(\mathbf x_1,\ldots,\mathbf x_n)$ and 
$I_1=(\mathbf y_1,\ldots,\mathbf y_n)$,
results in the momentum equation
\begin{align}
  Lu_t
  =
  -
  \frac{1}{\lambda^2} 
  \sum_{i=1}^n
  Dg_{t,1}(\mathbf x_i(1))^{-T} ( \mathbf x_i(1)- \mathbf y_i)
  \delta_{\mathbf q_i(t)}\, , 
  \label{momentum-equation-landmarks}
\end{align}
with landmark position $\mathbf x_i(t)=g_{t,0}(\mathbf x_i)$ at time $t$.

\subsection{Iterative Matching: The Beg Algorithm}

The algorithm for LDDMM image matching presented in \cite{beg2005computing}
performs a gradient descent optimization to fulfil the momentum equation
\eqref{momentum-eq}. Expressed using the momentum map, the gradient 
$\nabla_{u} E(u)$ with respect to the $V$-norm takes the form
\begin{align}
  \nabla_{u} E(u)
  =
  2u_t
  -
  K\left (
  \frac{2}{\lambda^2} J_t^0\diamond \left ( g_{t,1}(J_1^0-J_1^1)^\flat \right)
  \right ).
\end{align}
This equation holds, in general, for all shape data types.
The gradient descent algorithm updates $u_t$ iteratively as
\begin{align}
  u_t^{k+1}
  =
  u_t^k
  -
  \epsilon
  \nabla_{u} E(u^k)\, , \quad \forall t\, . 
  \label{beg-updates}
\end{align}
The algorithm can be interpreted as a gradient flow by introducing
an additional time parameter $s\in\mathbb R^+$ in which case \eqref{beg-updates} arise as a
discretized version of the flow 
\begin{align}
  \partial_s u_{t,s}
  =
  -
  \epsilon
  \nabla_{u} E(u)_{t,s}
  \label{beg-flow}
  \, .
\end{align}
We will see the time parameter $s$ appearing again in the string methods
in Section~\ref{sec:sto-inexact-match}. The actual numerical algorithm presented 
in \cite{beg2005computing}
includes a reparametrization step after each $k$ ensuring the velocity fields
$u_t$ are of unit speed.

\subsection{Exact Matching}

The case of exact matching can be treated as a variational boundary value problem
without the dissimilarity term of \eqref{E-functional}, by formally setting
$\lambda=0$. The action integral, in this case, contains only kinetic energy.
Instead of specifying
that the flow must satisfy the reconstruction equation \eqref{rec-det}, we can
instead, introduce an advection condition by adding a corresponding term directly to the 
variation formulation. This results in the action integral
\begin{align}
  S(u,p,I) 
  = 
  \int_0^1 l( u_t) dt 
  + 
  \int_0^1 \pair{p_t,\partial_tI+\mathsterling_{u_t}I}_Vdt
  \,.
  \label{S-functional}
\end{align}
The dual elements $p\in V^*$ act as Lagrange multipliers ensuring that the dynamic
variable, here the shape $I$, is advected by the flow. i.e. for optimal $(u,p,I)$,
$\partial_tI+\mathsterling_{u}I=0$ for all $t$. In the case of exact matching,
the momentum map is given by
\begin{align}
  \frac{\delta l}{\delta u}
  =
  I\diamond p
  \, .
  \label{mom-eq-exact}
\end{align}
In contrast to the momentum map in \eqref{momentum-eq},  
the momentum map in this case is independent of the initial and target shapes. 
However, although the momentum map changes when passing
to the exact matching case, the dependence on the endpoints in the inexact case
disappears when taking time derivatives and the EPDiff equation \eqref{EP} is
the same for both exact and inexact matching.

\subsection{Momentum Map based Shape Stochastics}

The importance of the derivation of the deterministic dynamics in
terms of the momentum map is that the stochastic shape model
introduced in \cite{arnaudon_geometric_2018} preserves 
geometric structure; in particular, it preserves the
momentum map. This is achieved by introducing stochasticity in the
reconstruction equation \eqref{rec-det} via its stochastic equivalent
\begin{align}
  dg_t = u_t g_t dt + \sum_{i=l}^J \sigma_l g_t \circ dW^l_t\,,
    \label{rec-sto}
\end{align}
corresponding to the stochastically perturbed flow vector field, $d\tilde{u}_t$, 
\begin{align}
  d\tilde u_t= u_tdt + \sum_{i=l}^J \sigma_l \circ dW^l_t
  \, .
  \label{tilde-u}
\end{align}
Compared to \eqref{rec-det}, \eqref{rec-sto} has an additional finite sum of
$J$ fields $\sigma_l\in\mathfrak g$ multiplied by the coordinate increments of a 
$J$-dimensional Brownian motion $W_t\in\RR^J$ with standard filtrations $\mathcal F_t^l$, see for example \cite{oksendal_stochastic_2003} for more details. The stochastic derivative is
defined using Stratonovich integration $\circ$. We note that while the
stochastic perturbation is here finite dimensional, the model can be extended to
infinite dimensional noise as in e.g. \cite{vialard2013extension}.

The energy functional \eqref{E-functional} remains unchanged, except that
$g_1$ is found as a solution to the perturbed reconstruction equation
\eqref{rec-sto}. We write $E(\tilde{u};I_0,I_1)$ to emphasize this, and
reserve $E(u;I_0,I_1)$ for deterministic $u$ with the reconstruction
\eqref{rec-det}.
Notice that the paths are only non-smooth with respect to the time variable $t$, but they remain smooth with respect to the space variables. 
It can now be proved by direct calculation that the momentum map equation
\eqref{momentum-eq} is unchanged by the stochastic perturbation of $dg_t$. 
By taking time derivatives of the momentum equation \eqref{momentum-eq}, 
the following result for general shape spaces is derived in \cite{arnaudon_geometric_2018}:
\begin{proposition}
  \label{prop:sto-shape}
  With the stochastically perturbed reconstruction equation \eqref{rec-sto},
  the momentum equation \eqref{momentum-eq} is unchanged, and
  a path being critical for \eqref{E-functional}, i.e. $u$ satisfies
  $\nabla_{u}E(\tilde{u})=0$,
  is equivalent to $u_t$ satisfying the stochastic Euler-Poincar\'e equation
\begin{align}
    d \frac{\delta l}{\delta u} + \ad^*_{u_t} \frac{\delta l}{\delta u}dt+
    \sum_{l=1}^J \ad^*_{\sigma_l} \frac{\delta l}{\delta u}\circ dW^i_t=0\, ,  
    \label{EP-sto}
\end{align}
\end{proposition}
Note that the critical paths of \eqref{E-functional} depend on the noise realization. The proposition gives necessary equations for $u_t$ to be optimal for each fixed noise realization. Consequently, the optimal $u_t$ are random variables.

The presence of noise in the reconstruction equation was
first introduced in \cite{holm2015variational}. The term `parametric
stochastic deformation' emphasises that the spatial dependence of
solutions is parametric and only the temporal
dependence is stochastic, see also discussions in \cite{HoTy2016}. 
The fields $\sigma_1,\ldots,\sigma_J\in\mathfrak g$
can be considered a spatial basis for the noise, and the spatial correlation between
the perturbations is controlled by $\sigma_l$. With sufficient smoothness on
$\sigma_l$ and $l$ sufficiently strong, flows with finite energy will be almost 
surely diffeomorphic. The parameters for the fields $\sigma_l$ can be inferred from data by solving an inverse problem, see below or \cite{arnaudon_geometric_2018}.

Because the momentum map is preserved for the perturbed flows,
the stochasticity descends to any of the shape spaces on which the diffeomorphism
group acts.
As in the deterministic setting, the fact that the momentum map takes different 
forms depending on the infinitesimal action of $\mathfrak g$ on the shape space
results in different dynamics for the different type of shapes.

Following \cite{holm2015variational} and \cite{HoTy2016}, the
corresponding stochastic version of the exact matching action functional is
\begin{align}
  S(\tilde{u},p,I)
  =
  \int_0^1
  l(u_t)dt
  +
  \int_0^1
  \left<p,dq+\tilde{\mathsterling}_{\tilde{u}_t}I
  \right>
  dt
  \, .
  \label{S}
\end{align}
where $\tilde{\mathsterling}_{u_t}$ is a stochastic Lie differential 
that for general vector valued quantifies $v$ takes the form
\begin{align}
  \tilde{\mathsterling}_{\tilde{u}_t} v
  =
  \mathsterling_{u_t} vdt
  -
  \sum_{l=1}^J
  \mathsterling_{\sigma_l} v\circ dW_t^l\, , 
  \label{sto-Lie-diff}
\end{align}
using the regular Lie derivative $\mathsterling$. Notice that the value of this Lie
derivative is a stochastic integral.
As in the inexact matching case, the momentum equation is unaffected by the
stochastic perturbation. Using the stochastic Lie derivative, the stochastic EPDiff 
equation can be written
\begin{align}
  d\frac{\delta l}{\delta u}
  =
  \tilde{\ad}_{d\tilde{u}}\frac{\delta l}{\delta u}\, , 
  \label{sto-epdiff-exact}
\end{align}
with $\tilde{\ad}_{\tilde u}\xi=-[\tilde\mathsterling_{\tilde
u},\mathsterling_\xi]$, or, equivalently,
\begin{align}
  dm
  +
  \tilde\mathsterling_{d \tilde u}m
  =
  0\, , 
  \label{sto-epdiff-exact-lie}
\end{align}
with $m=\frac{\delta l}{\delta u}$.
These stochastic equations are considered in the
landmark case in \cite{arnaudon_geometric_2018} leading to the finite dimensional stochastic differential equation (SDE)
\begin{align}
    \begin{split}
        d \mathbf q_i &= \sum_j \mathbf p_jK(\mathbf q_i-\mathbf q_j) dt + \sum_l\sigma_l(\mathbf q_i) \circ d W_t^l\\
        d \mathbf p_i &= 
        -\sum_j \mathbf p_i\cdot \mathbf p_j\partial_{\mathbf q_i}K(\mathbf q_i-\mathbf q_j)\ dt 
        - \sum_l \partial_{\mathbf q_i}\left (\mathbf p_i \cdot\sigma_l(\mathbf q_i)\right )  \circ  d W_t^l \, .
    \end{split}
    \label{sto-Ham-explicit}
\end{align}
The equations extend the usual deterministic LDDMM landmark equations by added
Stratonovich perturbation terms that are dependent on the fields $\sigma_l$.

\section{Stochastic Image Dynamics and Inexact Matching} \label{sec:sto-images}

We now aim at specializing the general stochastic dynamics as surveyed in
Section~\ref{sec:background} to the case of images to get the dynamic image
equations in the stochastic case, and to later extend the Beg algorithm as
originally presented for images to the stochastic setting.

For sufficiently smooth images $I:\Omega\rightarrow\RR$, the momentum field
$m(\cdot,t)$ will be a spatially differentiable 1-form density. In coordinates for
$\Omega\subseteq\RR^d$, we write $m(x)=\mathbf m(x)\cdot d\mathbf x\otimes d^dx$. The
deterministic EPDiff equation in the image case is then often written in
coordinates as
\begin{align}
  \partial_t \mathbf m
  +\mathbf m\cdot \nabla \mathbf u
  +\mathbf u\cdot \nabla \mathbf m
  +\mathrm{div}(\mathbf u)\mathbf m
  =
  0
  \, , \quad 
  \partial_t I
  =
  -\mathbf u\cdot\nabla I
  \label{det-image-epdiff}
  \, .
\end{align}
This form arises from the fact that the coadjoint action $\ad^*$ for 1-form densities
equals the Lie derivative so that \eqref{EP} takes the form
\begin{align}
  \partial_tm+\mathsterling_{u}m=0 \, ,
    \label{EP-Lie}
\end{align}
with $m=\frac{\delta l}{\delta u}$, and by computing the Lie derivative
$\mathsterling_{u}m$ in coordinates. 

In the stochastic exact matching case,
we saw above that \eqref{EP-Lie} generalizes to stochastic dynamics using the
stochastic differential
\eqref{sto-epdiff-exact-lie} and the stochastic Lie differential
\eqref{sto-Lie-diff}. 
The stochastic addition to the EPDiff equation comes from the left-most term of
\eqref{sto-Lie-diff}. Computing the Lie derivatives $\mathsterling_{\sigma_l} v$
gives the following vector form of the stochastic term
\begin{align}
  \sum_{l=1}^J
  \big(
  (\mathbf \sigma_l\cdot \nabla) \mathbf m
  +\mathbf m\cdot (D\mathbf \sigma_l)^T
  +\divergence(\mathbf \sigma_l)\mathbf m
  \big)
  \circ
  dW^l_t
  \, .
  \label{sto-addition-image-epdiff}
\end{align}
Combined with the deterministic part, this gives the stochastic image EPDiff equation 
\eqref{sto-vector-EPDiff-intro} in the exact matching case.
The image evolution that in the deterministic setting follows the usual advection equation
$\partial_t I=-u\cdot\nabla I$ becomes the stochastic integral
$dI+\tilde\mathsterling_{\tilde u}I=0$.

Turning to inexact matching, because the momentum equation is preserved by
Proposition~\ref{prop:sto-shape},
the image momentum equation \eqref{momentum-equation-images} still holds for the deterministic part $u_t$ of
$\tilde u_t$. As in the deterministic case, the fact that the momentum map is
different between the exact and inexact matching case does not affect the
dynamic equations. Using Proposition~\ref{prop:sto-shape} and calculating the
coordinate expressions of the stochastic EPDiff equation \eqref{EP-sto} as for
\eqref{sto-addition-image-epdiff}, we arrive at the following vector version of
the inexact image matching stochastic EPDiff equation that generalizes the
deterministic equation \eqref{det-image-epdiff}
\begin{align}
  \begin{split}
  &
  d \mathbf m
  +
  \big(
  (\mathbf u\cdot \nabla) \mathbf m
  +\mathbf m\cdot (D\mathbf v)^T
  +\divergence(\mathbf u)\mathbf m
  \big)dt
  \\
  &\qquad
  +
  \sum_{l=1}^J
  \big(
  (\mathbf \sigma_l\cdot \nabla) \mathbf m
  +\mathbf m\cdot (D\mathbf \sigma_l)^T
  +\divergence(\mathbf \sigma_l)\mathbf m
  \big)
  \circ
  dW^l_t
  = 0
  \\
  &
  dI=
  -\mathsterling_{u_t} Idt
  +
  \sum_{l=1}^J
  \mathsterling_{\sigma_l} I\circ dW_t^l
  =
  -\nabla I\cdot \mathbf udt
  +
  \sum_{l=1}^J
  \nabla I\cdot\mathbf\sigma_l\circ dW_t^l
  \, .
  \end{split}
  \label{vector-image-inexact-epdiff}
\end{align}
Notice the interaction between the image gradient and the noise fields in the
stochastic advection equation for the image $I$.

\section{The Stochastic Beg Algorithm and String Methods} \label{sec:sto-inexact-match}

The momentum representation leads to a direct generalization of the matching
algorithm \eqref{beg-updates} to the stochastic setting. 
As for the interpretation of the Beg algorithm as a discretized gradient flow,
we use an extra independent variable $s\in \RR^+$, 
a second time variable or time for the evolution of the curve $g_t$. For each $s$, the
flow will itself still be parametrised by the original time $t$.  
The momentum $m=\frac{\delta l}{\delta u}$ and thus $u$ and $g$ will now depend on 
the variable $s$, and we write the equation of motion for the $s$ evolution of
$m_{t,s}$ 
as
\begin{align}
    \begin{split}
      \partial_s m_{t,s} &= - \nabla_{u_{t,s}} E(\tilde{u})\\
    &= -m_{t,s}- \frac{1}{\lambda^2} J_t^0\diamond ( g_{t,1}(J_1^0-J_1^1)^\flat )\, .
    \end{split}
    \label{grad-descent}
\end{align}
analogous to \eqref{beg-flow} but here with $\nabla_{u_{t,s}} E(\tilde{u})$ taken with
respect to the $L^2$ pairing on $V^*$. When discretized in
the second time variable $s$, this gives a gradient descent like algorithm analogous 
to \eqref{beg-updates}.

In the deterministic setting, as $s\to \infty$, the system will converge to the stationary state  
corresponding to the equation \eqref{momentum-eq} and giving a solution of the matching problem. 
This extends to the stochastic setting with fixed noise $W^l_t$. Although the noise is not directly 
visible in \eqref{grad-descent}, it affects the
system via the reconstruction of $g_t$ given by the stochastically perturbed
reconstruction equation \eqref{rec-sto}, and because $g_t$ appears in the momentum equation. 

Below, we give different perspectives on the matching algorithm and flow
\eqref{grad-descent}, both as a string method and by comparing to an Expectation-Maximization
algorithm for finding the most probable curve between $I_0$ and $I_1$. After
this, we specialize the flow to the image and landmark cases.

\subsection{The String Method}

The string method developed in \cite{e_string_2002} without noise and
extended to include noisy strings in  \cite{e_finite_2005} is used for sampling
rare transition events and finding pathways in transition state theory. Analyzing phase 
transitions in physical systems is often complicated by the difference
between short time scales of the dynamics and much longer time scales of
transitions between metastable states, states in local minima of the energy
landscape. Monte Carlo simulations of the short time
dynamics thus have a low probability of giving information about the transitions
between states that are of interest. 

The string method was developed to solve this problem by sampling strings
between metastable states directly. In \cite{e_string_2002}, a string $g_t$
between states $g_0$ and $g_1$ is evolved according to
\begin{align}
  \partial_s
  g_{t,s}^\perp
  =
  -\nabla E(g_{t,s})^\perp\, , 
  \label{string}
\end{align}
where $\perp$ denotes the part of the $s$-derivatives point-wise orthogonal to the
$t$-derivative $\dot{g}_t$. The aim is to find a minimal energy path (MEP)
defined as a critical point of the energy, i.e.
\begin{align}
  \nabla_{ g_{t,s}}E(g)^\perp
  =
  0
  \label{MEP}
\end{align}
The projection $\perp$ ensures that the parameterization of the
string does not affect the dynamics. In practice, an arc length parametrization
can be chosen in which case the string is evolved for a fixed number of iterations before 
a reparametrization step enforces the arc length constraint.

In \cite{e_finite_2005}, the string method is
extended by adding finite temperature noise to the system resulting in the
addition of a noise term $\alpha\eta_{t,s}^\perp$ to \eqref{string} with $\alpha>0$ denoting the
finite temperature and $\eta_{t,s}$ t-dependent white noise along the string and parametrised by $s$, 
thus the noise affects both parameters $t$ and $s$.

Both string methods allow identification of MEPs between the starting and
ending states. The finite temperature sampling also allows estimation of transition
tubes along the MEP. The finite temperature method can be invoked with $M$ evolving strings, 
allowing the evolving MEP to be approximated by the average
\begin{align}
  g_{t,s}
  =
  \frac{1}{M}\sum_{j=1}^M
  g_{t,s}^j
  \, .
  \label{string-avg}
\end{align}
This gives information about the large scale effect of the
energy landscape on the dynamics. In particular, it can often happen that 
high-frequency features of the energy landscape have little effect on the
transition dynamics that to a higher degree are influenced by larger scale, low
frequency features such as energy barriers. As the temperature approaches zero, 
the finite temperature MEP approaches the MEP \eqref{MEP} of the original string method.
With non-zero temperature, the MEP should be seen as a generalized and averaged
equivalent of the MEP satisfying \eqref{MEP}.

In the present context, the image of the diffeomorphism flow $g_t$ acting on
$I_0$, i.e. $\{g_t.I_0\,|\,t\in[0,T]\}$ can be interpreted as a string from
$I_0$ to $g_T.I_0$ and the deterministic Beg algorithm with reparametrization
corresponds to the string equation \eqref{string}. 
In this context, the notion of rare event used for the original 
string method is slightly different. Indeed, in the standard application of 
the string method, the phase space is large, but of low dimension and the landscape
is irregular, with many local minima. In our case, the landscape is rather smooth, 
as mostly given by the kinetic energy, but the dimension of the phase space is large. 
In the case of $N$ landmarks, the string evolves in a $2dN$ dimensional space. 
Consequently, only rarely would a stochastic path emerging from a set of landmarks  
 reach another set of landmarks while solving the stochastic 
EPDiff equation. The string gives a notion of the average trajectory taken to achieve 
the random matching. One can also interpret the double well potential example 
of the string method where the pass is an obstacle to linking the two wells
as a large kinetic energy about half-way between the initial and target shapes. 
Indeed, the kinetic energy in this case plays the main role for the evolution of the string. 

In the stochastic setting,
allowing the noise to vary with $s$ gives an equivalent of the finite
temperature string method with noisy strings \cite{e_finite_2005}. In the shape case,
the noise does not appear directly as an additive term to the update equation
\eqref{string} but rather indirectly through the reconstruction equation. 
Our model is thus a nonlinear extension of the original finite 
temperature string method, with a particular type of multiplicative noise that preserves 
the structure of the equation. 
The original concept of the finite temperature string method persists in
this setting, but the analysis of the string sampling is harder; in particular, ergodicity
properties cannot be established directly. However, as we will see, we can
sample around MEPs equivalently to the string method and derive various
statistical information from shape string sampling.

A main feature of the string method is its computational efficiency. Since each
string update scales linearly in the number $n_t$ of discretization points in
the time $t$, $M$ strings can be evolved in $O(Mn_t)$. This evolution
parallelizes completely over several processing units.
In addition, in order to speed up convergence, the gradient descent flow \eqref{string}
is in \cite{e_string_2002} extended to a quasi-second order flow using a limited
memory method of Broyden's method. The flow is conditioned by a matrix that
approximates the inverse Hessian of the energy, and the convergence rate is
highly improved.

\begin{remark}
The inexactness of the matching is in \eqref{E-functional} measured at the
string endpoint. As discussed in \cite{bruveris2011momentum}, there are
various ways of symmetrizing the matching problem (in the sense of having both end images contributing the same to the matching term). 
One approach to make the energy symmetric is to measure the inexactness at both ends of the matching
\cite{hart_optimal_2009}
\begin{align}
  E_{\mathrm{sym}}(u_t,I,I_0,I_1) = 
  \int_0^1 l( u_t) dt 
  + \frac{1}{2\lambda^2} \|I - I_0\|^2
  + \frac{1}{2\lambda^2} \|g_1.I - I_1\|^2\, .
    \label{E-functional-sym}
\end{align}
This results in the momentum equation
\begin{align}
    \frac{\delta l}{\delta u_t}
    = - \frac{1}{\lambda^2} J_t\diamond ( g_{t,1}(J_1-J_1^1)^\flat )\, ,
    \label{momentum-eq-sym}
\end{align}
with $I=(I_0^\flat-g_{0,1}^{-1}(J_1-J_1^1)^\flat)^\sharp$
and $J_t=g_{t,0}.I$. The shape $I$ can be seen as a weighted average between
$I_0$ and $I_1$ mapped to $t=0$.
Because the momentum map is preserved in the stochastic scheme, symmetric
stochastic shape string
algorithms can be implemented analogously to the non-symmetric
algorithms.
\end{remark}

\subsection{Expectation-Maximization and Principal Curves} \label{sec:EM}

We can compare the string equation \eqref{grad-descent} to a stochastic Expectation-Maximization procedure
\cite{dempster_maximum_1977,delyon_convergence_1999} by interpreting the matching energy
\eqref{E-functional} as a negative $\log$-posterior density.
We assume 
the observed data $I=g_1.I_0+\epsilon$ is i.i.d. Gaussian distributed given the endpoint $g_1.I_0$. The complete data is now the deterministic part of the flow $u_t$, the noise process $W_t$, and $\epsilon$, however only
$I=g_1.I_0+\epsilon$ is observed. We define the incomplete data likelihood
\begin{align}
  \begin{split}
  &g(I|u_t)
  =
  \mathbb E[
  p(I|W_t,u_t)]\, , 
  \end{split}
  \label{eq:likelihood}
\end{align}
with Gaussian density for the image $I$ given $W_t,u_t$ and hence flow $\tilde{u}_t$:
\begin{align*}
  p(I|W_t,u_t)
  \propto
  \exp(-\frac{1}{2\lambda^2} \|g_1.I_0-I\|^2)
  \,  .
\end{align*}
In the image case, $V$ is infinite dimensional, and the density should be interpreted formally for a finite discretization of $V$, 
e.g., for a finite number of image pixels.

We consider $u_t$ a parameter for the model and, given an observed image $I_1$, we search for a maximum a posteriori estimate
\begin{align}
  \hat{u}_{\text{MAP}}
  \in
  \argmax_{u_t}
  p^{\text{flow}}(u_t)g(I_1|u_t)
  \, ,
  \label{eq:max_a_posteriori}
\end{align}
with prior $p^{\text{flow}}(u_t)\propto\exp(-\int_0^1l(u_t)dt)$ for the flow.
Notice that 
\begin{align*}
  -\log (p^{\text{flow}}(u_t)p(I_1|W_t,u_t))=E(\tilde{u}_t)+c
  \, .
\end{align*} 
The resulting model is analogous to the mixture
models used when identifying principal curves \cite{hastie_principal_1989} with
maxima of a corresponding likelihood function \cite{tibshirani_principal_1992}.
We refer to \cite{vanden-eijnden_revisiting_2009} for more details of the connection with principal curves, MEP and string methods. 
We now take a similar route to estimate maximally likely strings using the
EM-algorithm.

In the EM-algorithm, a maxima $\hat{u}_{\mathrm{MAP}}$ is found iteratively by
alternating the steps
\begin{description}
  \item[\bf E-step:] Compute (or estimate)
    \begin{align}
      \begin{split}
      Q(u_t|u_t^k)
      &=
      \mathbb E[\log (p^{\text{flow}}(u_t)p(I_1|W_t,u_t))|g_1.I_0+\epsilon=I_1]
      \\&
      =
      \mathbb E[-E(\tilde{u}_t)|g_1.I_0+\epsilon=I_1]
      - c
      \label{E-step}
      \, .
      \end{split}
    \end{align}
  \item[\bf M-step:]
    Increase (or maximize) $Q$ wrt. $u_t$:
    \begin{align}
      \begin{split}
      u^{k+1}
      &=
      u^k
      +
      \epsilon
      \nabla_{u_t}
      Q(u_t|u_t^k)
      =
      u^k
      -
      \epsilon
      \mathbb E[
      \nabla
      E(\tilde{u}_t)|g_1.I_0+\epsilon=I_1]\, . 
      \end{split}
      \label{M-step}
    \end{align}
\end{description}
In the M-step, the expected negative gradient $\mathbb E[\nabla E(\tilde{u}_t)|g_1.I_0+\epsilon=I_1]$
given the current value $u_t^k$ of the string can be approximated by drawing a
finite number of samples, evaluating $\nabla E(\tilde{u}_t)$, i.e. the right hand side
of the string
equation \eqref{grad-descent}, and reweighting by $p(I_1|W_t,u_t^k)/g(I_1|u_t^k)$. 
The minimizer of the stochastic matching functional \eqref{E-functional}, the string
MEPs, and curves $\hat{u}_{\mathrm{MAP}}$ under the model \eqref{eq:max_a_posteriori} thus differ in this reweighting in the expectation, or, equivalently, in the expectation in \eqref{E-step} being conditional on $I_1$.

When the variance $\lambda^2$ of $\epsilon$ is small and the scheme is relatively close to exact matching, the filtering provided by $p(I_1|W_t,u_t^k)/g(I_1|u_t^k)$ in the expectation will generally lead to many low-probability samples. A dedicated bridge sampling approach is developed in \cite{arnaudon_geometric_2018} for the landmark case to sample directly from the data conditional distribution and alleviate this problem. With larger $\lambda^2$, the filtering is less pronounced and the need for dedicated sampling schemes reduced. The string method does not have the filtering term and thus computes the expectation unconditional on the observed data while still taking gradients of the $\log$-posterior of $u_t$ given the observed data.

\subsection{String Method for Landmarks}

For the numerical experiments given in Section~\ref{sec:experiments}, we here insert
the landmark action in the string equation \eqref{grad-descent} to derive the string evolution for
stochastic landmarks explicitly. Using the momentum map
\begin{align}
    \mathbf m(x,t,s)= \sum_{i=0}^N \mathbf p_i(t,s) \delta_\mathbf x(\mathbf q_i(t,s))\, , 
    \label{momap}
\end{align}
the equation \eqref{grad-descent} for landmarks simplifies to
\begin{align}
    \partial_s \mathbf p_i(t,s) &=  -\mathbf p_i(t,s)- \frac{1}{\lambda^2}
    Dg_{t,1}(\mathbf q_i(1))^{-T} ( \mathbf q_i(1)- \mathbf q_i(t))\, ,
    \label{string-landmark}
\end{align}
where 
\begin{align}
    d \mathbf q_i(t)= u_t(\mathbf q_i(t))) dt + \sum_l \sigma_l(\mathbf q_i(t)) \circ dW_t^l(s)\, , 
    \label{phi_eq}
\end{align}
and
\begin{align}
    u_t(x) =  \sum_i K(x-\mathbf q_i(t))) \mathbf p_i(t)\, .
    \label{ux}
\end{align}
We refer for example to \cite{arnaudon_geometric_2018} for more details on the derivation of these equations.

The matrix $Dg_{t,1}(q_i(1))$ is computed by differentiating \eqref{phi_eq}
(using $q_i^\alpha(t)= g_t(q_i(0))^\alpha$) to get  the backward in time equation
\begin{align}
  \begin{split}
    &d Dg_{t,1}(\mathbf q_i(1))^{\alpha,\beta} = - D u_t( \mathbf q_i(t)))^\alpha_\gamma
    Dg_{t,1}(\mathbf q_i(1))^{\gamma,\beta} dt \\
    &
    \qquad\qquad
    \qquad\qquad
    \qquad
    + \sum_l D \sigma_l( q_i(t)))^\alpha_\gamma Dg_{t,1}(q_i(1))^{\gamma,\beta}
    \circ dW_t^l(s)\, , 
  \end{split}
    \label{Dphi}
\end{align}
where
\begin{align}
    D u_t(x)^\alpha_\gamma=  \sum_i \partial_{x^\gamma} K(x-\mathbf q_i(t))) p_i^\alpha(t)\, , 
\end{align}
with initial condition $Dg_{1,1}(q_i(1))=\mathrm{Id}$. 
The processes $W_t(s)$ are standard Weiner processes in the $t$ variable. For the zero-temperature string method, the noise is not dependent on $s$. For the finite-temperature string method, the noise is a Wiener process in the $s$ variable for each fixed $t$ as well.

The string method has an extra feature, namely the projection of the right-hand side of \eqref{grad-descent} and the direction perpendicular to the string (in the $t$) variable, that is equation \eqref{string}.
We will not apply this projection here as it is used to allow reparametrisation of
the string for the more difficult matching problems in rare event sampling. 
We refer to \cite{beg2005computing} for a reparametrisation procedure in the context of image matching. 
For landmark matching, the reparametrisation can take place in the $\mathbf q_i$
variables and the $\mathbf p_i$ variables must be updated accordingly so that the approximation of the continuous string remains the same after the reparametrisation. 

The numerical scheme for a sequence $s_k$, initial conditions $\mathbf p_i(0,s_0)$, and
$\mathbf q_i(0,s)$, is displayed in Algorithm~\ref{alg:string} for constant temperature,
i.e. optimization to convergence for each noise realization,
and in Algorithm~\ref{alg:finite-string} for finite temperature, i.e. new noise
realization for each iteration of the algorithm.
\begin{algorithm} \DontPrintSemicolon \SetAlgoLined
  draw noise realization $\omega$\\
  \For{$k=1$ to $n_s$}{ 
    given $\mathbf p_i(t,s_k)$ for all $t$, compute $\mathbf q_i(t,s_k)$ from \eqref{phi_eq} and \eqref{ux} \\
    compute $Dg_{t,s_k}$ from \eqref{Dphi}\\
    compute $\mathbf p_i(t,s_{k+1})$ from \eqref{string-landmark}
  }
  \caption{Stochastic Beg algorithm: Landmark strings, constant temperature.}
  \label{alg:string}
\end{algorithm}
\begin{algorithm} \DontPrintSemicolon \SetAlgoLined
  \For{$k=1$ to $n_s$}{ 
    draw noise realization $\omega_k$\\
    given $\mathbf p_i(t,s_k)$ for all $t$, compute $\mathbf q_i(t,s_k)$ from \eqref{phi_eq} and \eqref{ux} \\
    compute $Dg_{t,s_k}$ from \eqref{Dphi}\\
    compute $\mathbf p_i(t,s_{k+1})$ from \eqref{string-landmark}
  }
  \caption{Stochastic Beg algorithm: Landmark strings, finite temperature.}
  \label{alg:finite-string}
\end{algorithm}

\subsection{Strings Method for Images}

Using the image momentum map and \eqref{grad-descent}, the image string update equation is, together with the reconstruction relation \eqref{rec-sto}, given by
\begin{align}
  \partial_s u(t,s)
  =
  -2u_t
  +K\left (
  \frac{2}{\lambda^2} 
  \left|\det Dg_{t,1}^{-1}\right| (J_t^0-J_t^1)\nabla J_t^0
  \right )\, . 
  \label{eq:image-string}
\end{align}
The projection happens as a
reparametrisation of the string after each step, similarly to the reparametrisation in the original Beg
matching algorithm. With fixed noise, the discretized string evolution is
identical to the Beg matching algorithm with the only difference being
the perturbed reconstruction equation. With finite temperature, the algorithms differ
only in that new noise is drawn for each update of $s$.

\section{Statistical Analysis of Matching} \label{sec:stat}

Given i.i.d. shape observations $I^1,\ldots,I^n$, we here give examples of how
the string method can be used for statistics of the observations and for 
parameters inference in the model.

\subsection{Mean Strings}

A mean string can be defined as
\begin{align}
  \bar u_t
  =
  \frac{1}{n}\sum_{i=1}^n
  \mathbb E_{\tilde{u}_t|u_t}
  [
    \argmin_{u_t}\,E(\tilde{u}_t;I_0,I^i)
  ]\, , 
  \label{eq:zero-mean-string}
\end{align}
which can be approximated by iterating the zero temperature string method to
convergence for each $i=1,\ldots,n$ and for each sampled noise realization.
The finite temperature equivalent arises via sampling new noise for each iteration.

\subsection{Frech\'et Mean Estimation}

The Frech\'et mean \cite{frechet_les_1948} on a Riemannian manifold $M$ of a distribution
$X$ is defined as a minimizer of the expected square distance to $X$, i.e.
\begin{align}
  \argmin_x \mathbb E[d_M(x,X)^2]
  \, .
\label{F-mean}
\end{align}
The distance $d_M(x,y)$ here denotes the geodesic distance between two points $x$ and $y$.

Though the inexact matching energy \eqref{E-functional} is not a square distance, we can
nevertheless, define a sample average of the observations that resembles the Frech\'et mean by
\begin{align}
  \bar I = \argmin_I\sum_{i=1}^n\min_{u}E(u;I,I^i)\, .
\end{align}
In the stochastic setting, $\bar I$ will be a random variable depending on
perturbations in the reconstruction equation. We can define the zero temperature average
\begin{align}
  \bar I = \argmin_I\sum_{i=1}^n
  E_{\tilde{u}_t|u_t}[\min_{u}\mathbb E(\tilde{u};I,I^i)]\, , 
  \label{eq:FM}
\end{align}
as well as its finite temperature equivalent by drawing new noise for each iteration of gradient descent iterative optimization of \eqref{eq:FM}.

\subsection{Parameter Inference}

We can consider any combination of $I$, parameters of the kernel $K$, and the
noise fields $\sigma_1,\ldots,\sigma_J$, unknowns of the model and seek to estimate these unknowns from the observations $I^1,\ldots,I^n$. A direct approach 
is to compare statistics of the observations with statistics of the distribution arising at the string endpoints, either with zero or finite temperature. For observed landmark configurations $\mathbf q^1,\ldots,\mathbf q^n$, this can be sample mean and covariance of each landmark $\mathbf q^i_j$ compared with sample mean and covariances of the string endpoint landmark configurations. The method of 
moments is used in \cite{arnaudon_geometric_2018} in a similar fashion for landmark 
parameter inference, although by direct approximation of the landmark density function
instead of string sampling.

\section{Numerical Experiments} \label{sec:experiments}

We here present examples of matching with the string method and finite
temperature string method in addition to estimation of the expected 
mean. The experiments are performed on landmarks and image manifolds with 
the LDDMM metric. The code for performing the experiments is available in the repository
\url{http://bitbucket.com/stefansommer/stochlandyn}. See also \cite{kuhnel_differential_2017} for more info on the use of automatic differentiation frameworks for differential geometry computations.

In the landmark case, we use both synthetic data and
points representing the shape of left ventricles in cardiac images.
The noise fields are kernels of the form 
\begin{align}
    \sigma_l^\alpha (\mathbf q_i)
    = \lambda_l^\alpha  k_{r_l}(\|\mathbf q_i-\delta_l\|)\,,
    \label{kernels}
\end{align}
with noise amplitude $\lambda_l\in \mathbb R^d$, length scale $r_l$ and with 
$\delta_l$ denoting the kernel positions. Here $k_{r_l}$ is a Gaussian
$k_{r_l}(x)=e^{-\|x\|^2/(2r_l^2)}$.
For the LDDMM kernel, we similarly use Gaussian kernels. For all experiments performed 
here, 16 noise kernels
$\sigma_1,\ldots,\sigma_{16}$ with fixed length scale and amplitude are placed in the 
shape domain on a regular 4x4 grid.
Since the kernel is not of compact support, kernel multiplications such as in the forward flow \eqref{Dphi} scale quadratically in the number of evaluation points. Each iteration of the string method thus scales linearly in the number of evaluation points $n_t$ and quadratically in the number of evaluation points, i.e. the number of landmarks $N$.
\begin{figure}[hpb]
\centering
\includegraphics[width = 0.48\textwidth, trim = 150 80 110 80, clip]{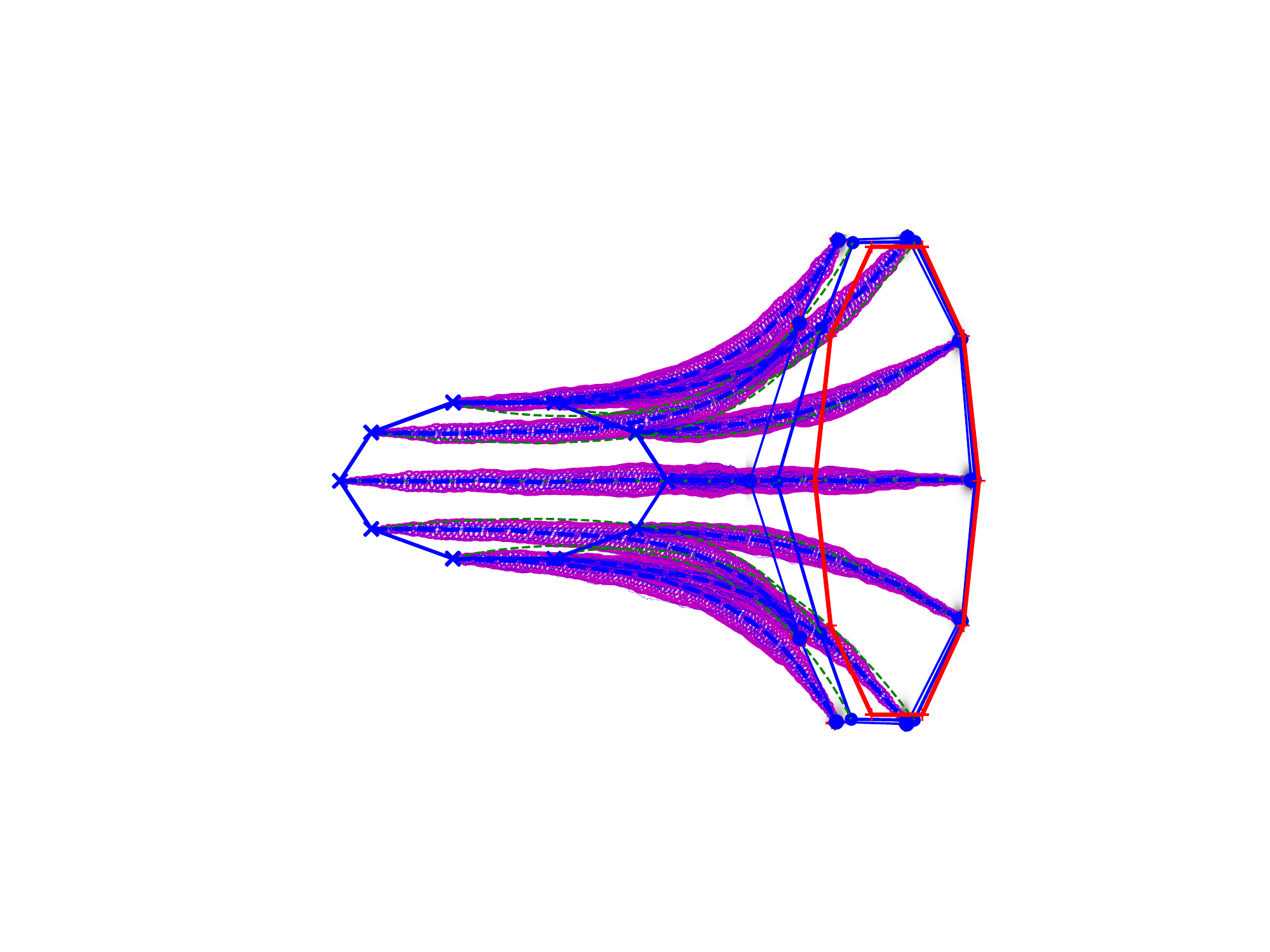}
\includegraphics[width = 0.48\textwidth, trim = 150 80 110 80, clip]{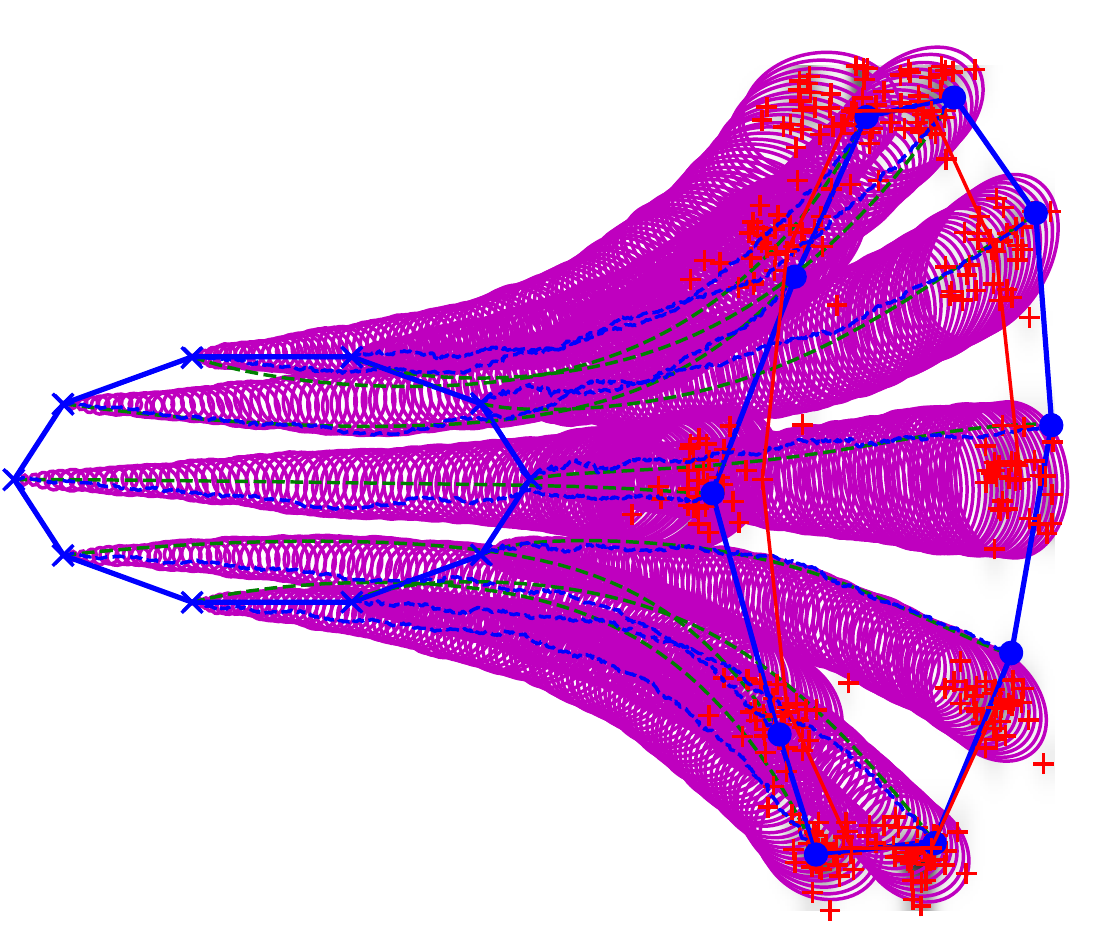}
\caption{Landmark strings matching $I_0$ (blue solid
  lines/crosses) towards a target $I_1$ (red solid line). Left:
  zero temperature, right: finite temperature. Samples from the endpoint
distribution (red crosses)
and the finite energy mean string (fat blue dashed). For each $t$ and landmark
$\mathbf q_{i,t}$, covariance of the samples $\mathbf q_{i,t,s_k}$ (ellipses)
show the effect of the noise perturbations.}
\label{fig:string2}
\end{figure}
\begin{figure}[hpb]
\centering
\includegraphics[width = 0.4\textwidth, trim = 150 60 110 60, clip]{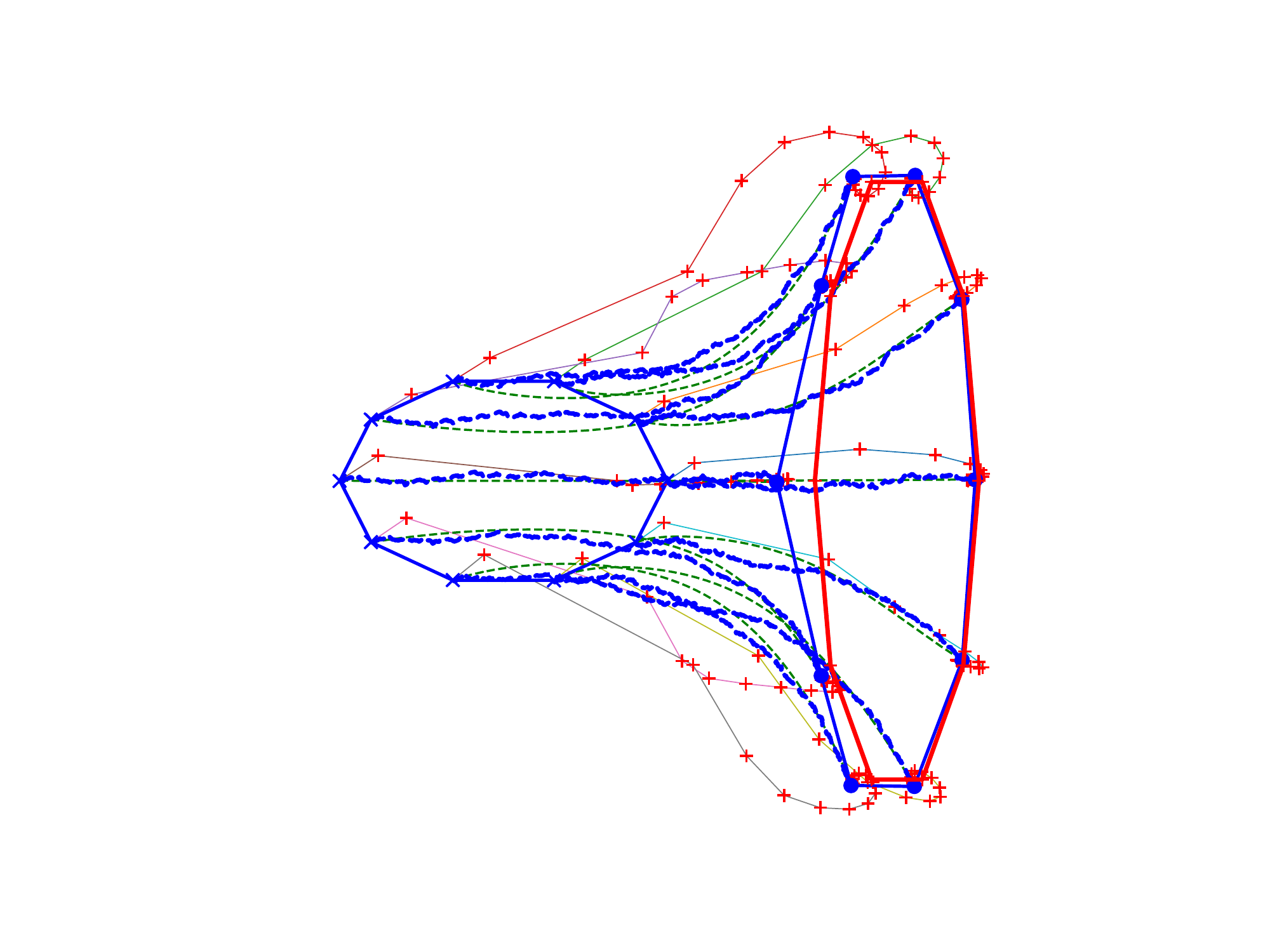}
\caption{ Convergence of the zero temperature string method. String
  endpoint configuration for each $s_k$ (red +) shown for all $k$.
Because the same noise realization is used with zero temperature, 
the algorithm smoothly moves the string from the initial configuration and converges towards the target as $s$ increases. The final path (blue
dashed) appear as a perturbation of the deterministic optimal path (green dashed). 
}
\label{fig:string_zero1}
\end{figure}

\subsection{Synthetic Data Landmark Data}

With the setup as Figure~\ref{fig:string1}, we arrange 10 landmarks in two ellipse configurations.
We first run the string method (Figure~\ref{fig:string2}, left) and finite temperature string method
(Figure~\ref{fig:string2}, right). Samples from the endpoint distributions are shown along with
estimated MEP $\hat{\mathbf q}_t$ and sample covariance of $\mathbf q_{t,s^k}$.
The MEP can be compared to the minimizing geodesic between the landmark configurations.
The string is at $s=0$ with zero velocity, i.e. $\mathbf q_{t,0}=\mathbf q_{0,0}$ for all $t$.
The fact that non-zero temperature increases the variance of the string 
and that the sample covariance increases with time $t$ is
clearly visible.
\begin{figure}[tpb]
\centering
\includegraphics[width = 0.48\textwidth, trim = 40 40 90 40, clip]{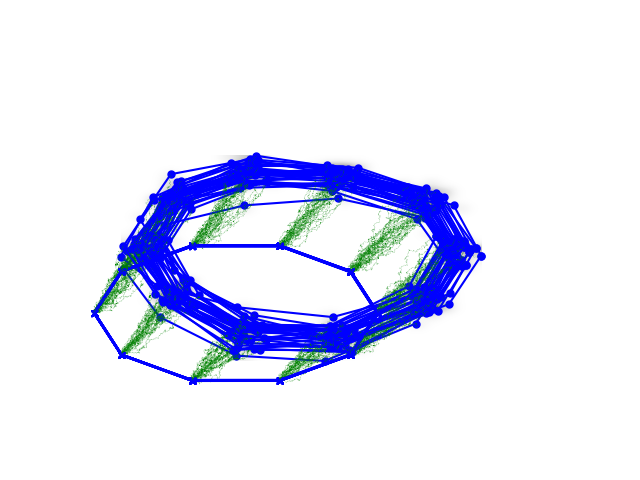}
\caption{Sample landmark configurations (blue solid lines) generated by sampling
  from the endpoint configuration of the perturbed landmark EPDiff equation.
  Green lines show perturbed trajectories to the samples from the initial
configuration.}
\label{fig:FM_samples}
\end{figure}
\begin{figure}[hpb]
\centering
\includegraphics[width = 0.44\textwidth, trim = 70 80 130 80, clip]{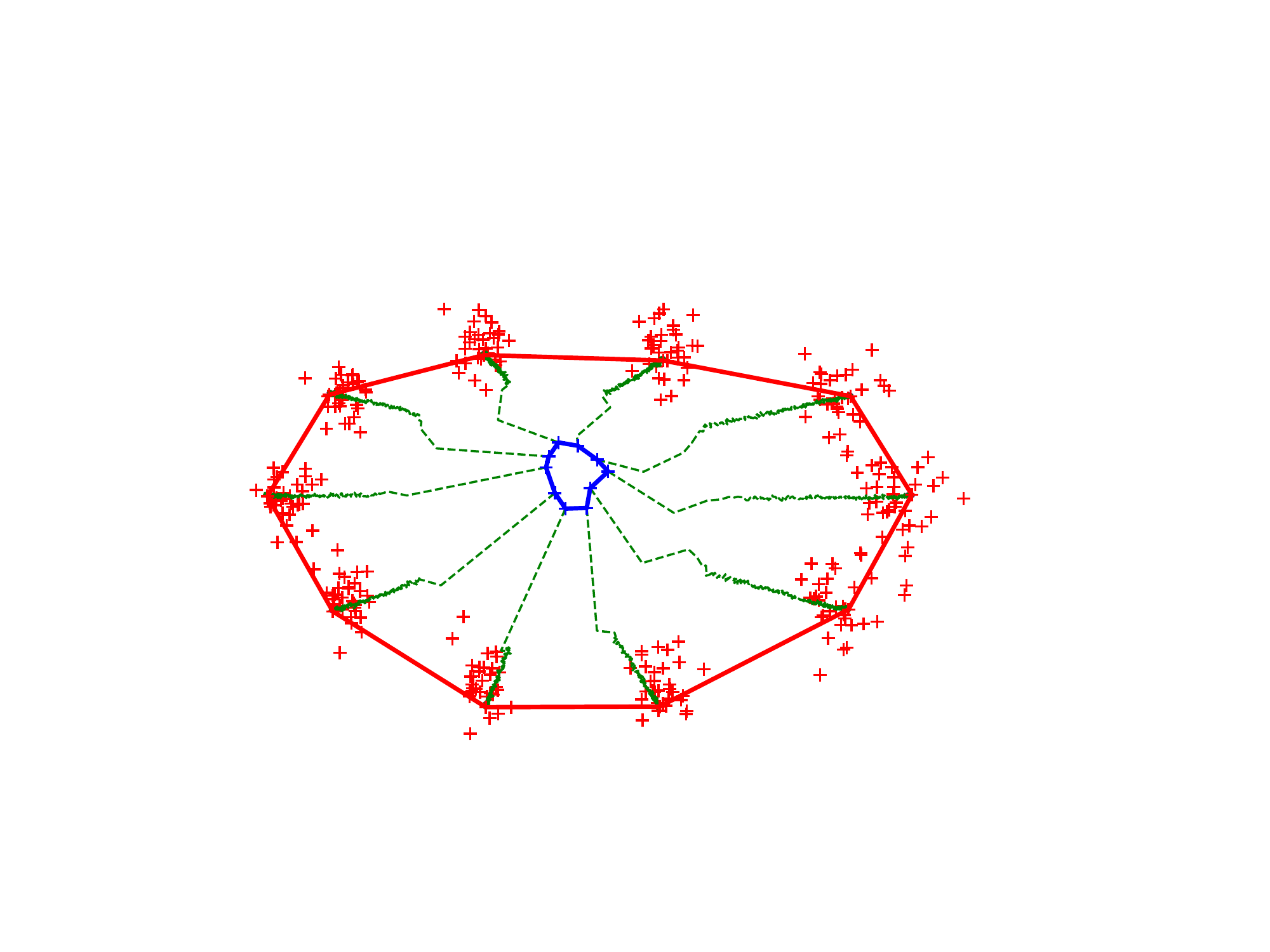}
\includegraphics[width = 0.44\textwidth, trim = 70 80 130 80, clip]{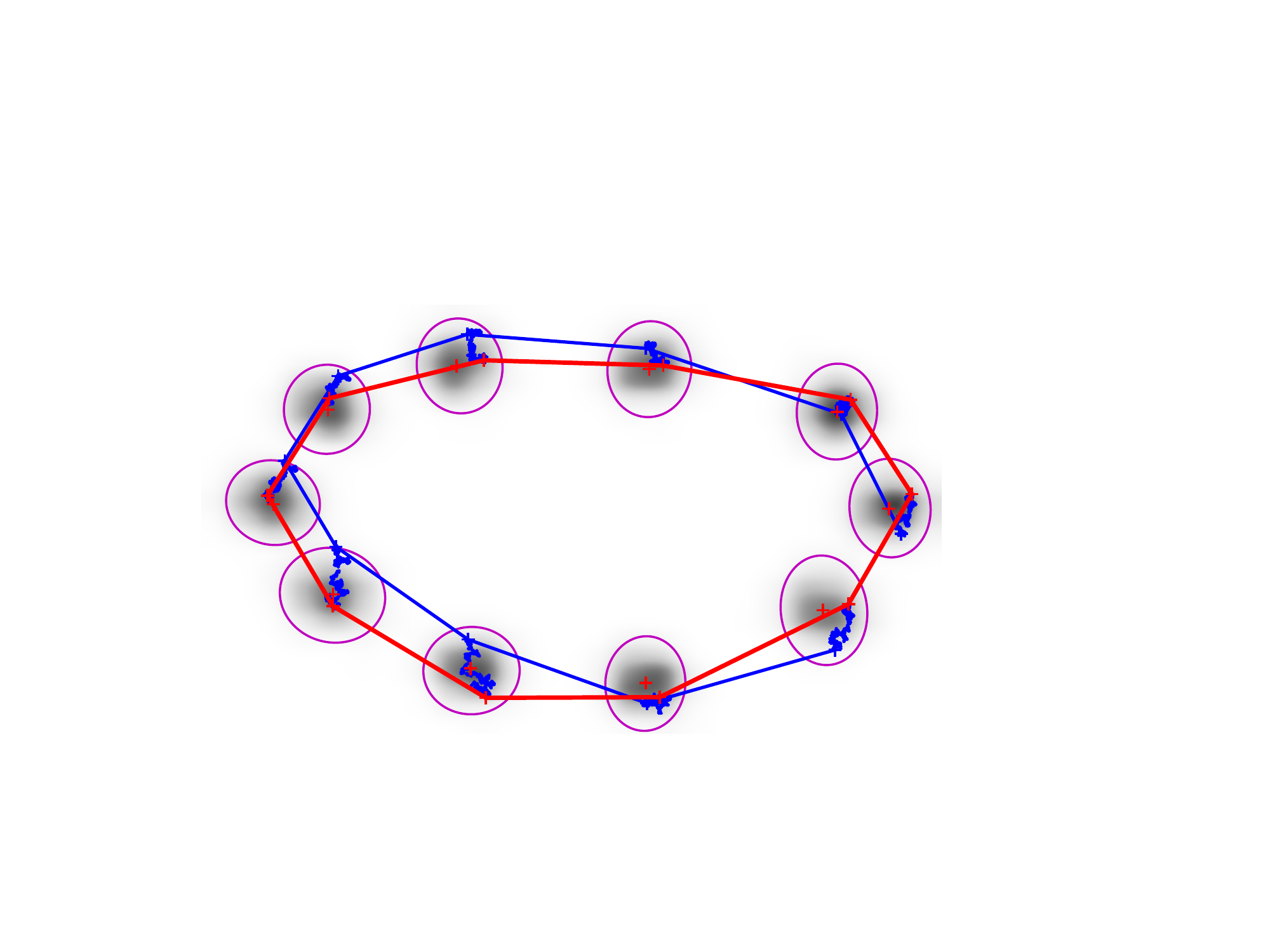}
\caption{(left) Using samples from Figure~\ref{fig:FM_samples}, evolution (green
dashed) of the mean landmark configuration from random initial value
  (blue) toward the estimated mean (red). (right) String from the estimated mean
  (red) to one of
the samples (blue) together with samples from the endpoint configuration matching the
mean to the sample.}
\label{fig:FM}
\end{figure}

Figure~\ref{fig:string_zero1} shows an example of the convergence of the string with zero
temperature, i.e. with single noise realization. The endpoint configuration
at $t=1$ converges smoothly as a function of $s$. The final converged string is a perturbed
version of the optimal deterministic string.

We now compute the expected mean \eqref{eq:FM} of a new set of sampled configurations shown in
Figure~\ref{fig:FM_samples}. The algorithm is initialized with a random
configuration, and the evolution of the expected mean configuration can be seen
in Figure~\ref{fig:FM} (left) together with a string matching the mean to a
sample (right). The mean converges to what visually appears a to a reasonable mean
landmark configuration.

\subsection{Left Ventricle Cardiac Outlines}

\begin{figure}[hpb]
\centering
\includegraphics[width = 0.48\textwidth, trim = 20 20 20 20, clip]{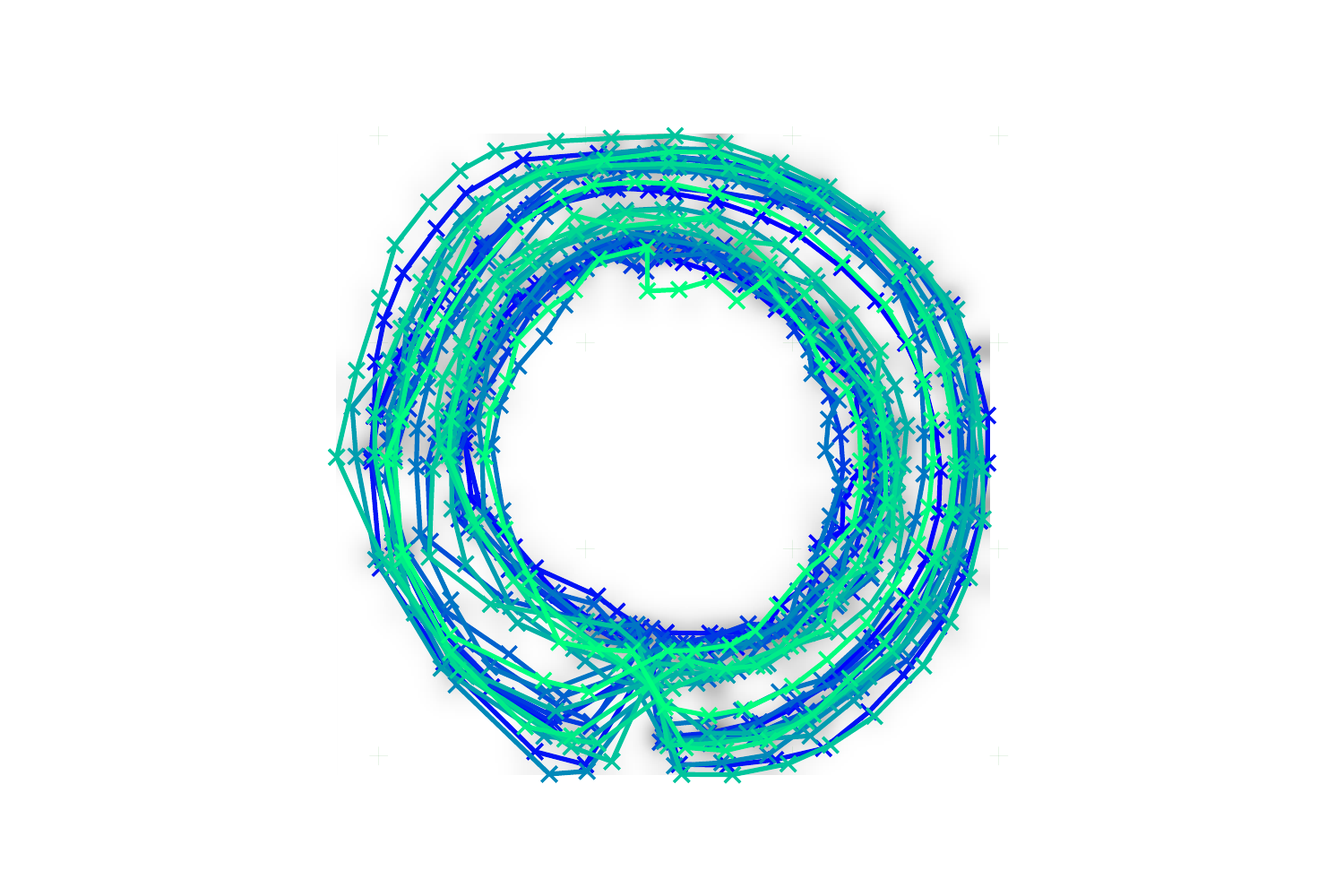}
\includegraphics[width = 0.40\textwidth, trim = 140 80 120 80, clip]{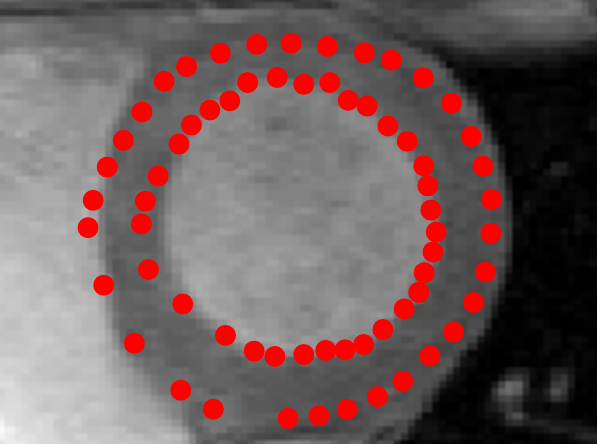}
\caption{(left) 14 landmark configurations obtained from the cardiac images.
(right) One cardiac image with landmark annotated left ventricle.}
\label{fig:cardiac1}
\end{figure}

\begin{figure}[hpb]
\centering
\includegraphics[width = 0.48\textwidth, trim = 70 60 90 60, clip]{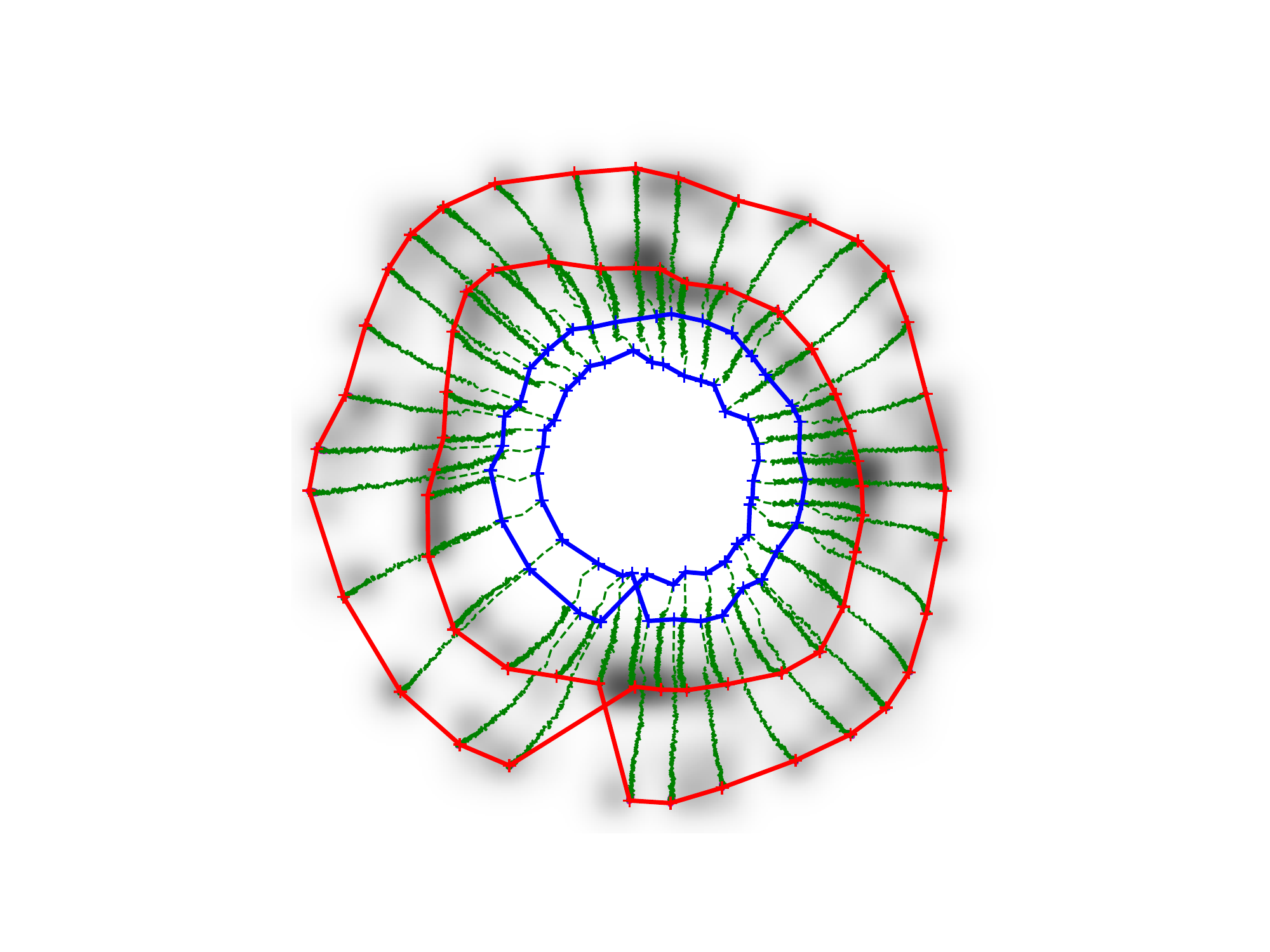}
\includegraphics[width = 0.48\textwidth, trim = 70 60 90 60, clip]{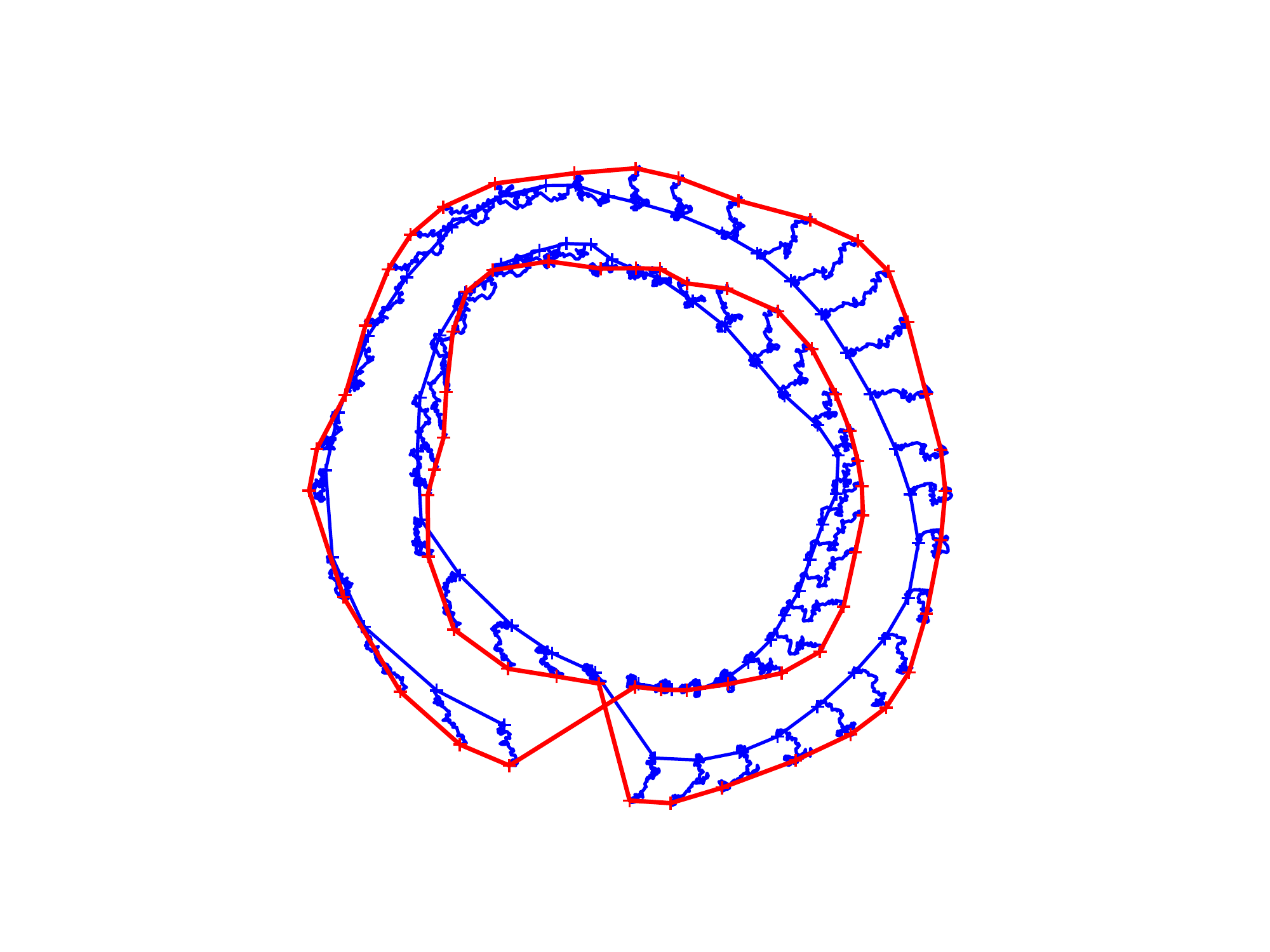}
\caption{(left) Evolution (dashed green) of the mean estimate of the ventricle annotations from
  initial configuration (blue) to estimated mean (red) overlayed density estimate of the mean
  estimates as a function of time $s$. The initial configuration is the Euclidean mean of the landmark configurations, rescaled and added i.i.d. noise.
(right) String from estimated mean (red) to one of the annotated ventricle
configurations (blue).}
\label{fig:cardiac2}
\end{figure}

To illustrate the method on non-synthetic data, we perform experiments on landmarks
distributed along the outlines of left ventricles on a dataset of 14 cardiac images
\cite{stegmann_extending_2001}.  Each of the 256$\times$256 MRI slices is acquired from 1.0 Tesla whole-body MR scans with ECG-triggered breath-hold sequences. The epicardial and endocardial contours where annotated with 33 landmarks along each outline resulting in 66 landmarks per image. The set of annotations are shown in
Figure~\ref{fig:cardiac1} together with an annotated image. With the higher number of landmarks per outline and the double circular configuration of the landmarks, the matching problem is more difficult than for the synthetic examples.

\begin{figure}[htpb]
\centering
\includegraphics[width = 0.48\textwidth, trim = 40 40 40 40, clip]{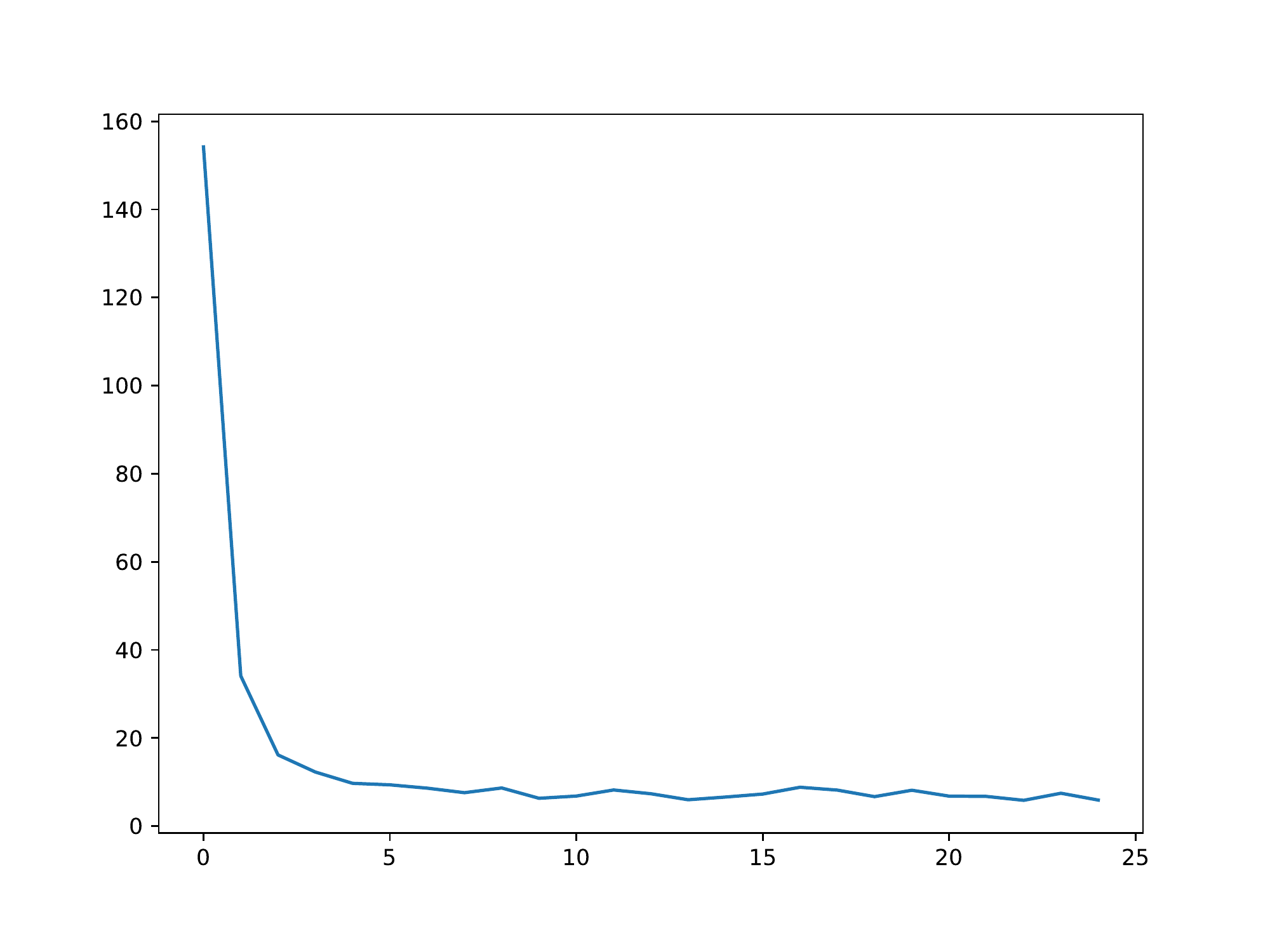}
\caption{The energy \eqref{eq:FM} with finite temperature for the first 25 iterations of the mean estimation for the ventricle annotations.
}
\label{fig:cardiac3}
\end{figure}

As in Figure~\ref{fig:FM}, Figure~\ref{fig:cardiac2} shows results of estimating the mean along with
a string connecting the estimated mean to one of the data samples. 
Initialized with
a configuration of landmarks in the centre of the image (blue), the mean 
converges in a stable way towards the final estimate (red).
The energy as a function of $s$ for the first 25 iterations is displayed Figure~\ref{fig:cardiac3}. Up to the stochasticity from the sampling, it converges monotonically from its initially high value as the landmarks approach the mean configuration.

\subsection{Image Strings}
We now use the image equation \eqref{eq:image-string} to provide an example of matching with stochastic image strings and the effect of the noise on the image evolution. We here use cubic B-spline kernels $k_{r_l}$ for the noise positioned in a $9\times9$ grid over the domain with amplitude $\gamma_l$ and length scale $r_l$ set to make the noise amplitude uniform over the domain. Image gradients in the stochastic $dI$ image flow in \eqref{vector-image-inexact-epdiff} are computed by finite differences, and the flow field $u_t$ arise from the momentum field $m_t$ by application of the kernel $K$, here again, a convolution with a Gaussian kernel.

\begin{figure}[htpb]
\centering
\includegraphics[width = 0.32\textwidth, trim = 80 30 80 30, clip]{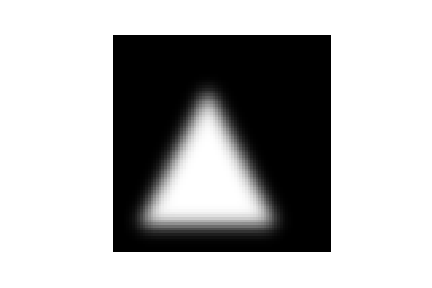}
\includegraphics[width = 0.32\textwidth, trim = 80 30 80 30, clip]{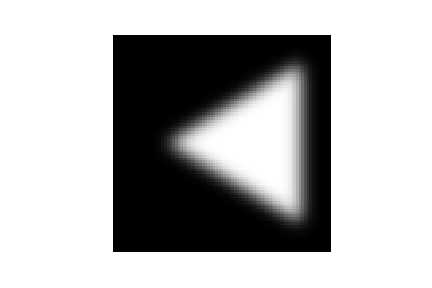}
\includegraphics[width = 0.32\textwidth, trim = 80 30 80 30, clip]{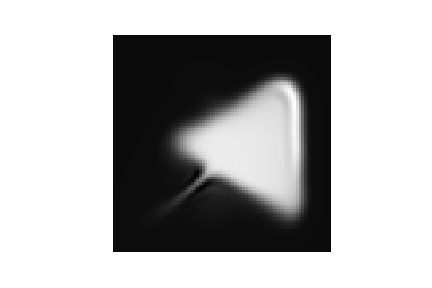}
\\
\includegraphics[width = 0.21\textwidth, trim = 80 30 80 30, clip]{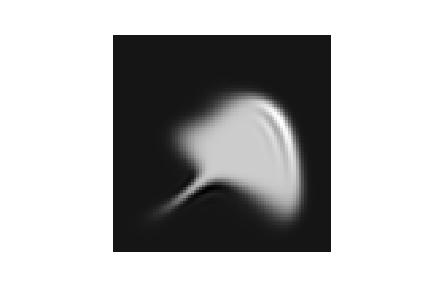}
\hspace{-.5cm}
\includegraphics[width = 0.21\textwidth, trim = 80 30 80 30, clip]{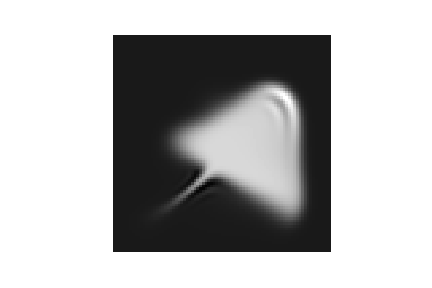}
\hspace{-.5cm}
\includegraphics[width = 0.21\textwidth, trim = 80 30 80 30, clip]{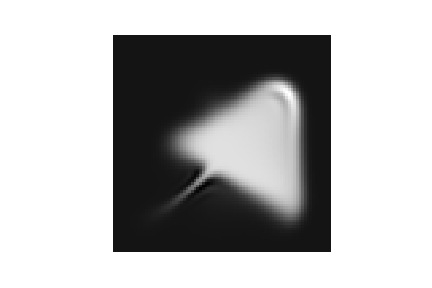}
\hspace{-.5cm}
\includegraphics[width = 0.21\textwidth, trim = 80 30 80 30, clip]{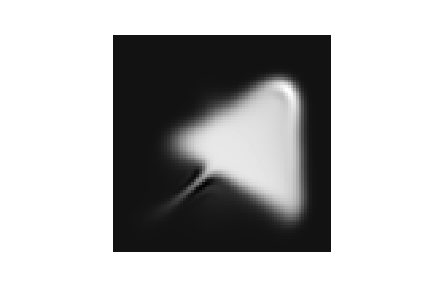}
\hspace{-.5cm}
\includegraphics[width = 0.21\textwidth, trim = 80 30 80 30, clip]{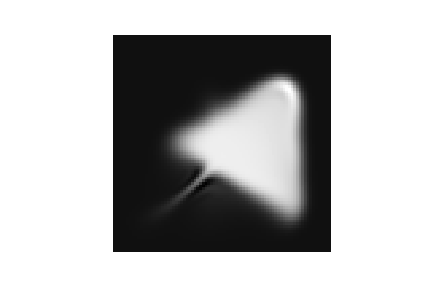}
\caption{Image matching with the string method. Top row, left: fixed image $I_0$. Center: target image $I_1$. Right: moving image $g_1.I_0$ after convergence of the algorithm. Bottom row: Moving image $g_1.I_0$ after 10, 30, 50, 70, 90 iterations ($s$) of the algorithm.}
\label{fig:image1}
\end{figure}

In Figure~\ref{fig:image1}, a triangle ($I_0$, top row, left) is matched to a triangle ($I_1$, top row, centre) with the stochastic algorithm giving the result in the top row, right. The bottom row shows the evolution of the moving image $g_1.I_0$ as a function of the second time variable $s$ during the iterations of the matching algorithm. The momentum field $m_t$ and hence the velocity field $u_t$ are initialized to zero at the start $s=0$ of the algorithm. The matching is inexact as can be seen by the triangle protrusion in the matching result that would require a higher warp energy to disappear fully.

\begin{figure}[htpb]
  \centering
\includegraphics[width=.99\columnwidth, trim = 150 30 120 30, clip]{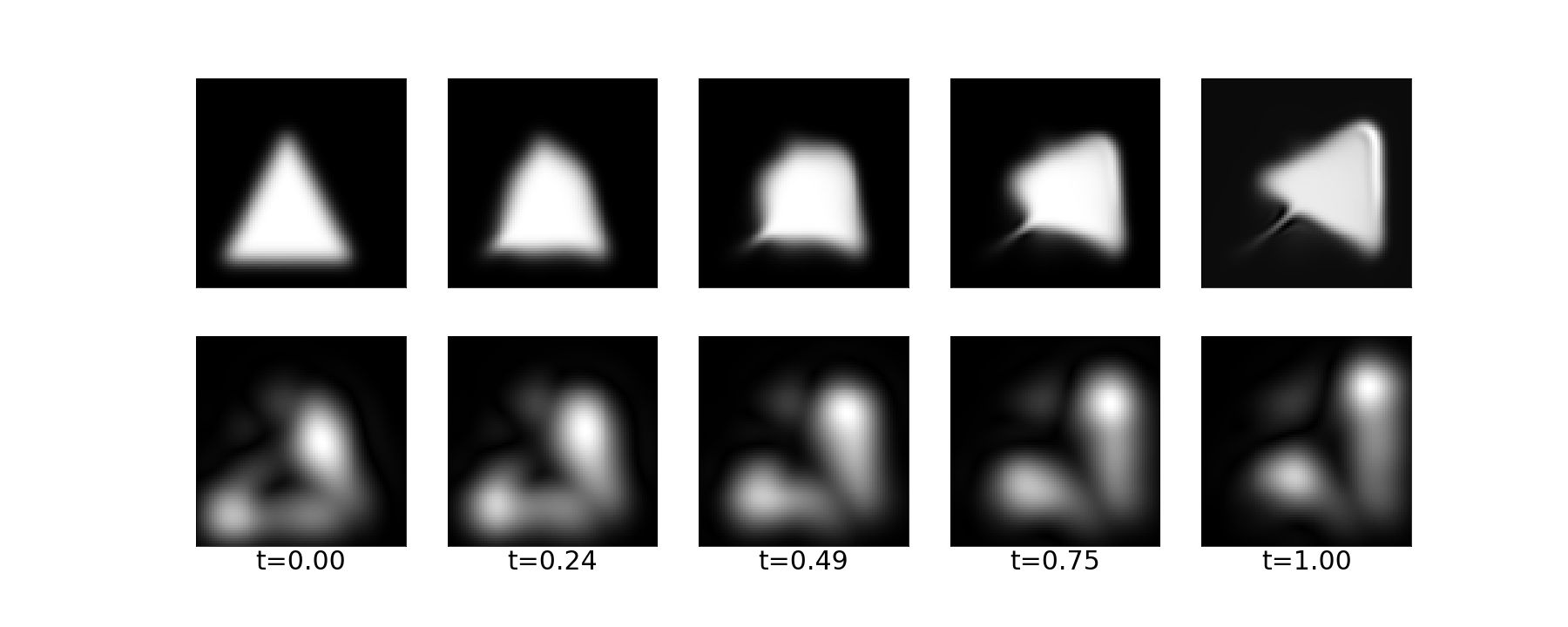}
\includegraphics[width=.99\columnwidth, trim = 150 30 120 80, clip]{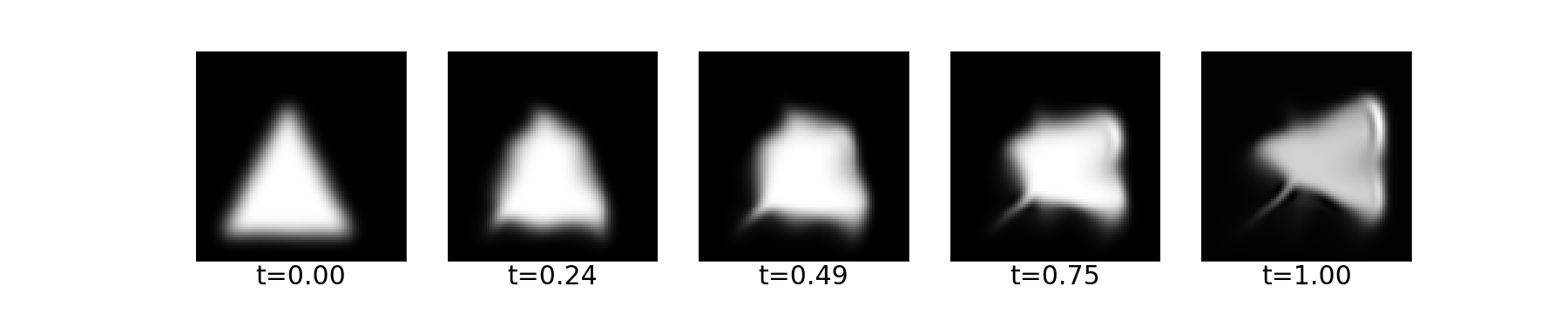}
\includegraphics[width=.99\columnwidth, trim = 150 30 120 30, clip]{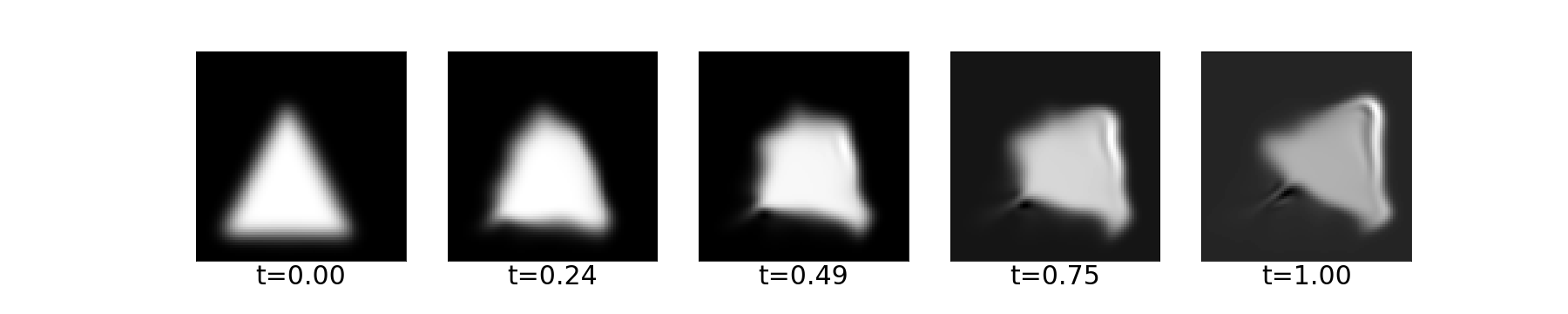}
\includegraphics[width=.99\columnwidth, trim = 150 30 120 80, clip]{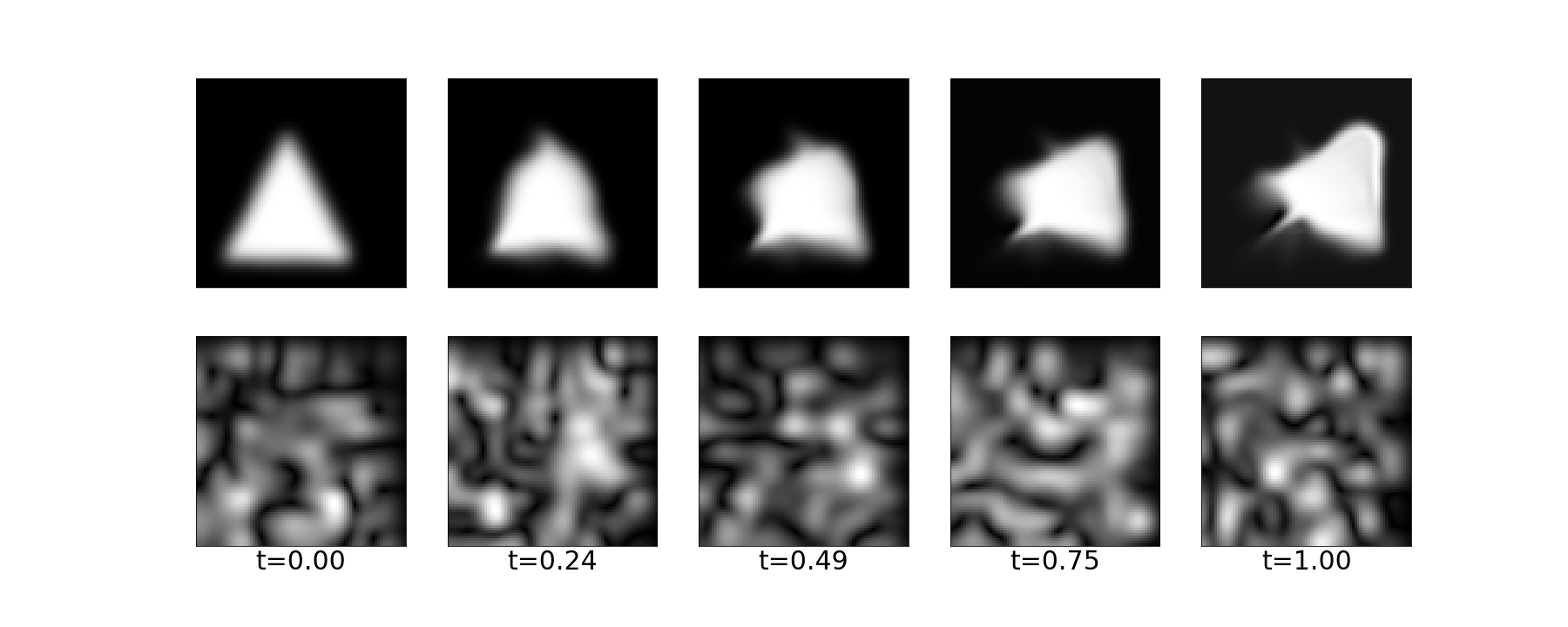}
\caption{Image string after convergence of the algorithm: Row 1: $g_t.I_0$ for $t=0,.24,.49,75,1$ with $dW=0$. Row 2: amplitude plot of the non-perturbed flow field $u_t$. Row 3-5: Image strings $g_t.I_0$ for the five values of $t$ and three different noise realizations. Row 6: Amplitude plot of the stochastically perturbed flow field $\tilde{u}_t$ corresponding to the image flow in row 5.}
\label{fig:image2}
\end{figure}

The algorithm runs with finite temperature drawing new noise for each iteration. Figure~\ref{fig:image2} illustrates the effect of the noise after the final iteration of the matching algorithm. The two top rows show the final image string with zero noise as a function of the first time variable $t$, and a magnitude plot of the corresponding velocity field $u_t$. Notice how the deformation is localized at the edges of the images. Row 3-5 display the image string as a function of $t$ for three different noise realizations. Row 6 shows the stochastically perturbed velocity field $\tilde{u}_t$ corresponding to the image flow in row 5. The amplitude is here not concentrated around the edges of the image in contrast to the situation in row 2. The effect of the stochastic perturbations to the flow is substantial. Note that the perturbations are changing during the iterations of the matching algorithm affecting the gradients of the matching terms. However, the averaging over the noise realizations provided by the string method makes the estimated deterministic trajectory stable to these perturbations as seen in the top row of the figure.

\section{Conclusion}

Shape stochasticity can be introduced to model stochastic shape variation 
in a way that is compatible with the geometric structure of the LDDMM framework.
In this setting, optimal dynamics arise from a matching energy that is dependent on the
stochastically perturbed reconstruction equation, or from a constrained and
the stochastically perturbed variational principle in the exact matching case.
In this paper, we derived the image case of the stochastic EPDiff equations for
inexact shape matching, and showed how they extend the vector form of the
deterministic EPDiff equation by addition of a Stratonovich perturbation term.

The matching algorithm used in deterministic LDDMM often referred to as the Beg
algorithm has a direct counterpart in the stochastic case because the noise is
introduced to preserve the momentum equation. We have shown how the stochastic Beg
algorithm is a shape equivalent of the string methods used in rare event
sampling. The shape string method can be used with both zero and finite temperature.

We provided examples of how the string method can be used for computing
statistics of observed data in a computationally efficient way, and we gave
examples of the shape string method and string based statistics on finite dimensional
landmark manifolds and images equipped with LDDMM geometry. The momentum map representation of shapes \cite{bruveris2011momentum} and the preservation of the momentum map in the stochastic setting allows the method to be applied to other shape data types beyond these examples. One such case would be matching of tensor fields as pursued in Diffusion Tensor MRI where examples of momentum maps for selected choices of actions are given in \cite{bruveris2011momentum}. As for the landmark and image equations, once the momentum map is established, the string update equations follow directly.

\subsubsection*{Acknowledgements}

{\small 
AA acknowledges partial support from an Imperial College London Roth Award and the EPSRC through award EP/N014529/1 funding the EPSRC Centre for Mathematics of Precision Healthcare.
AA and DH are partially supported by the European Research Council Advanced Grant 267382 FCCA held by DH. DH is also grateful for support from EPSRC Grant EP/N023781/1.
SS is partially supported by the CSGB Centre for Stochastic Geometry and Advanced Bioimaging funded by a grant from the Villum Foundation. 
We thank the anonymous reviewers for insightful comments that have substantially improved the manuscript.
}

\bibliographystyle{alpha}  
\bibliography{biblio}

\end{document}